\documentclass[11pt,a4paper]{article}
\usepackage{times,latexsym}
\usepackage{url}
\usepackage[T1]{fontenc}
\usepackage{booktabs}
\usepackage{amsmath}
\usepackage{amsfonts}
\usepackage{graphicx}
\usepackage{subcaption}
\usepackage{comment}

\usepackage[acceptedWithA]{tacl2018v2}

\usepackage{xspace,mfirstuc,tabulary}

\newif\iftaclinstructions
\taclinstructionsfalse 
\iftaclinstructions
\renewcommand{\confidential}{}
\renewcommand{\anonsubtext}{(No author info supplied here, for consistency with
TACL-submission anonymization requirements)}
\newcommand{\instr}
\fi

\iftaclpubformat 

\else

\fi

\newcommand{\tacl}[1]{\textcolor{red}{#1}}

\title{Dealing with Disagreements:\\ Looking Beyond the Majority Vote in  Subjective Annotations}

\author{Aida Mostafazadeh Davani \\
  University of Southern California \\
  \texttt{mostafaz@usc.edu} \\\And
  Mark Díaz \\
  Google Research \\
  \texttt{markdiaz@google.com} \\\And
  Vinodkumar Prabhakaran \\
  Google Research \\
  \texttt{vinodkpg@google.com} \\}

\date{}

\begin{document}
\maketitle
\begin{abstract}

Majority voting and averaging are common approaches employed to resolve annotator disagreements and derive single ground truth labels from multiple annotations. However, annotators may systematically disagree with one another, often reflecting their individual biases and values, especially in the case of subjective tasks such as detecting affect, aggression, and hate speech. Annotator disagreements may capture important nuances in such tasks that are often ignored while aggregating annotations to a single ground truth. In order to address this, we investigate the efficacy of multi-annotator models. In particular, our multi-task based approach treats predicting each annotators' judgements as separate subtasks, while sharing a common learned representation of the task. We show that this approach yields same or better performance than aggregating labels in the data prior to training across seven different binary classification tasks. Our approach also provides a way to estimate uncertainty in predictions, which we demonstrate better correlate with annotation disagreements than traditional methods. Being able to model uncertainty is especially useful in deployment scenarios where knowing when not to make a prediction is important.

\end{abstract}

\section{Introduction}
Obtaining multiple annotator judgements on the same data instances is a common practice in NLP in order to improve the quality of final labels \cite{snow-etal-2008-cheap,nowak2010reliable}.
In case of disagreements between annotations, they are often aggregated by majority voting, averaging \cite{sabou2014corpus}, or adjudicating by an `expert' \citep{waseem2016hateful}, to derive a singular  ground truth or gold label that is later used for training supervised machine learning models. 
However, in many subjective tasks, there often exists no such single ``right'' answer \citep{alm2011subjective} and enforcing a single ground truth sacrifices the valuable nuances embedded in annotator's assessments of the stimuli and their disagreements \cite{aroyo2013crowd,cheplygina2018crowd} 

Annotators' socio-demographic factors, moral values, and lived experiences often influence their interpretations, especially in subjective tasks such as identifying political stances \citep{luo-etal-2020-detecting}, sentiment \cite{diaz2018addressing}, and online abuse and hate speech \cite{cowan2003empathy,waseem2016you,patton2019annotating}.
For instance, \newcite{waseem2016you} found that feminist and anti-racist activists systematically disagree with crowd workers on their hate speech annotations. 
Similarly, annotators' political affiliation affect how they annotate the neutrality of political stances \citep{luo-etal-2020-detecting}. 
An adverse effect of majority vote in such cases is limiting representation of minority perspectives in data \cite{prabhakaran2021releasing}, potentially reinforcing societal disparities and harms. 

Another consequence of majority voting, when applied to subjective annotations, is that the resulting labels may not be internally consistent. For example, consider a scenario where a sentence in a hate-speech dataset is annotated by a set of annotators, the majority of whom consider a phrase in it to be offensive, yet another sentence with the same phrase is annotated by a different set of annotators, the majority of whom \textit{do not} find the phrase to be offensive. 
Upon majority vote, the first sentence would be labeled as hate speech and the second sentence would not, despite containing similar content. 
Such inconsistencies in the majority label will add noise to the learning step, while the systematicity in the individual annotations is lost.

Finally, majority vote and similar aggregation approaches assume that an annotator's judgements about different instances are independent from one another. However, as outlined above, annotators' decisions are often correlated, reflecting their subjective biases. Prior work has investigated Bayesian methods to account for such systematic differences between annotators \cite{paun2018comparing},
however, they approach this as an alternate means to derive a single ground truth label, thereby masking 
the degree to which annotators disagreed.

Our proposed solution is simple: we introduce multi-annotator architectures to preserve and model the internal consistency in each annotators' labels as well as their systematic disagreements with other annotators. We show that the multi-task framework \cite{liu2019multi} provides an efficient way to implement a multi-annotator architecture that captures the differences between individual annotators' perspectives using the subset of data instances they labeled, while also benefiting from the shared underlying layers fine-tuned for the task using the entire dataset.
Preserving different annotators' perspectives until prediction step provides better flexibility for downstream applications. In particular, we demonstrate that it provides better estimates for uncertainty in predictions. This will improve decision making in practice, for instance, to determine when not to make a prediction or when to recommend a manual review.

Our contributions in this paper are three-fold: 1) We develop an efficient multi-annotator strategy that matches or outperforms baseline models on seven different subjective tasks by preserving annotators' individual and collective perspectives throughout the training process;
2) We obtain an interpretable way to estimate model uncertainty that better correlates with annotator disagreements than traditional uncertainty estimates across all seven tasks.
3) We demonstrate that model uncertainty correlates with certain types of error, providing a useful signal to avoid erroneous predictions in real-world deployments.

\label{sec_intro}

\section{Literature Review}

Learning to recognize and interpret subjective language has a long history in NLP \cite{wiebe2004learning,alm2011subjective}. 
While all human judgments embed some degree of subjectivity, it is commonly agreed that certain NLP tasks tend to be more subjective in nature.
Examples of such relatively subjective tasks include sentiment analysis \cite{pang-lee-2004-sentimental,liu2010sentiment}, affect modeling \cite{alm2008affect,liu2003model}, emotion detection \cite{hirschberg2003experiments,mihalcea2006corpus}, and hate speech detection \cite{warner2012detecting}.
\newcite{alm2011subjective} argue that achieving a single \textit{real `ground truth'} is not possible, nor essential, in subjective tasks, and call for finding ways to model subjective interpretations of annotators, rather than seeking to reduce the variability in annotations.
While which NLP tasks count as subjective may be contested, we focus on two tasks that are markedly subjective in nature.

\vspace{-5px}
\subsection{Detecting Online Abuse}

NLP-aided approaches to detect abusive behavior online is an active research area \cite{schmidt2017survey,mishra2019tackling,corazza2020multilingual}. Researchers have developed typologies of online abuse \citep{waseem2017understanding}, constructed datasets annotated with different types of abusive language \citep{warner2012detecting,price2020six,vidgen2020learning}, and built NLP models to detect them efficiently \citep{davidson2017automated, mozafari2019bert}. 
%
Researchers have also expanded the focus to more subtle forms of abuse such as condescension and microaggressions \cite{breitfeller2019finding,jurgens-etal-2019-just}.

However, recent research has demonstrated that these models tend to reflect and propagate various societal biases, causing disparate harms to marginalized groups.
For instance, toxicity prediction models were shown to have biases towards mentions of certain identity terms \citep{dixon2018measuring}, specific named entities \citep{prabhakaran-etal-2019-perturbation}, and disabilities \citep{hutchinson2020social}. Similarly these models are shown to overestimate the prevalence of toxicity in African American Vernacular English \citep{sap2019risk, davidson2019racial,zhou2021challenges}.
Most of these studies demonstrate association biases present in data; for instance, \newcite{hutchinson2020social} show that discussions about mental illness are often associated with topics such as gun violence, homelessness, and drugs, likely the reason for the learned association of mental illness related terms with toxicity.
While whether a piece of text is hateful or not depends also on the context \cite{prabhakaran-etal-2020-online}, not much work investigated the human annotator biases present in the training labels, and how they impact downstream predictions.

\vspace{-5px}
\subsection{Detecting Emotions}

Detecting emotions from language has been a significant area of research in NLP for the past two decades \cite{liscombe2003classifying,aman2007identifying,desmet2013emotion,hirschberg2015advances,poria2019emotion}. Annotated datasets used for training emotion detection models vary across domains, and use different taxonomies of emotions. While several datasets \citep{strapparava2007semeval, buechel2017emobank} include a small set of labels representing the six Ekman emotions \citep{ekman1992argument} --- \textit{anger}, \textit{disgust}, \textit{fear}, \textit{joy}, \textit{sadness}, and \textit{surprise}),  or bipolar dimensions of affect --- \textit{arousal} and \textit{valence} \citep{russell2003core}, others such as \newcite{demszky2020goemotions} and \newcite{crowdflower} include a wider range of emotion labels according to the Plutchik emotion wheel \citep{plutchik1980general} or the complex semantic space of emotions \cite{cowen2019mapping}.
%
%
Perceiving emotions is a subjective task affected by various contextual factors, such as time, speaker, mood, personality, and culture \citep{mower2009interpreting}. Since aggregating annotations of emotion expressions loses such contextual nuances, some researchers provide a distributional representation of emotions
\citep{fayek2016modeling, ando2018soft}. 
Here, we use annotations for the six Ekman emotions present in the dataset released by \citet{demszky2020goemotions} to demonstrate how our multi-annotator approach can capture emotions in a dis-aggregated fashion.


\vspace{-5px}
\subsection{Annotation Disagreement}
Researchers have studied different sources of annotator disagreements. \citet{krippendorff2011agreement} argued that there are at least two types of disagreement in content coding: random variation, that comes as an unavoidable by-product of human coding, and systematic disagreement, that is influenced by features of the data or annotators. \citet{dumitrache2015crowdsourcing} identifies different sources of disagreement as (a) the clarity of an annotation label (i.e., task descriptions), (b) the ambiguity of the text, and (c) differences in workers. \citet{aroyo2013crowd} also studied inter-annotator disagreement in association with features of the input, showing that it reflects semantic ambiguity of the training instances. Textual features have been shown to predict annotators' disagreement in determining the meaning of ambiguous words \cite{alonso2015predicting}.  Acknowledging inter-annotator disagreement as an indicator of annotator differences, \citet{kairam2016parting} clustered crowd-workers based on their annotation behaviors, and proposed a method for interpreting annotation disagreements and its sources.

For highly subjective tasks such as hate speech and emotions detection, annotation disagreements can be rooted in the differing subjectivities and value systems of annotators. In these cases, annotators build a subjective social reality as a basis for social judgments and behaviors \citep{greifeneder2017social}, which explains their labeling procedure. For example, in interviews with annotators in an aggression labeling task, \newcite{patton2019annotating} found that expert annotators from communities discussed in gang-related tweets drew on their lived experience to produce different label judgements compared with graduate student researchers. 
Such annotators whose lived experiences bring important perspectives to the task would be
dramatically underrepresented on generic crowd work platforms and, by definition, would be outvoted in disagreements subject to majority vote. Majority vote also necessarily obfuscates differences among groups underrepresented in annotator pools, such as older adults who can exhibit views on aging distinct from crowd workers \cite{diaz2020biases}, the majority of whom tend to be younger \citep{ross2010crowdworkers}.

Some studies have 
proposed alternatives to majority voting when aggregating multiple annotations.
In early work, \newcite{dawid1979maximum} used the \textit{EM} algorithm to obtain maximum likelihood estimates of the ``true'' label to account for annotator errors.
\citet{de2012did} used the individual annotation distributions to predict areas of uncertainty in veridicality assessment. \newcite{hovy2013learning} proposed an approach based on item-response model that uses posterior entropy to choose which annotators are trustworthy. \citet{waterhouse2013pay} developed 
a pointwise mutual information metric to quantify the amount of information in an annotator’s judgment that can be used to estimate the ``correct'' label of an instance. 
%
\citet{gordon2021disagreement} explore multiple annotators judgements to disentangle stable opinions from noise by estimating intra-annotator consistency. 
All these approaches aim to obtain the ``correct'' label, accounting for erroneous or non-trustworthy annotators, whereas we focus on retaining the annotator disagreements through the modeling process.

A few studies have explored approaches for utilizing annotation disagreement during model training. \citet{prabhakaran-etal-2012-statistical} explored applying higher cost for errors made on unanimous annotations to decrease the penalty of mis-labeling inputs with higher disagreement. 
Similarly, \newcite{plank-etal-2014-learning} incorporated annotator disagreement into the loss function of a structured perceptron model for better predicting part-of-speech tags.
Our work also utilizes annotator disagreements rather than resolving them in the data stage; however, we use a multi-task architecture using a shared representation to model annotator disagreements, rather than using it in loss function. 
\citet{cohn-specia-2013-modelling} use a multi-task approach to model annotator differences in machine translation annotations. While they use a Gaussian Process approach, we use the multi-task approach on top of pre-trained language models \cite{liu2019multi}.
\citet{chou2019every} proposed an approach where they model individual annotators separately in an inner layer to improve the final prediction. In contrast, our method uses the multi-task architecture, and provides the additional ability to utilize multiple predictions during deployment, for instance, to measure uncertainty. \citet{fornaciari-etal-2021-beyond} also leveraged annotator disagreement using a multi-task model that adds an auxiliary task to predict the soft label distribution over annotator labels, which improves the performance even in less subjective tasks such as part-of-speech tagging. 
In contrast, our approach models several annotators' labels as multiple tasks and obtains their disagreement.


\vspace{-5px}
\subsection{Prediction Uncertainty}

Model uncertainty denotes the confidence of model predictions, which has specific applications in non-deterministic machine learning tasks. 
For instance, interpreting model outputs and its confidence is critical in autonomous vehicle driving, where wrong predictions are costly or harmful \cite{schwab2019cxplain}.
In subjective tasks, uncertainty embeds additional information that supports result interpretation \citep{ghandeharioun2019characterizing}. For example, the level of uncertainty 
could help determine when and how moderators take part in a human-in-the-loop content moderation \cite{chandrasekharan2019crossmod,liu2020human}.

The simplest approach for uncertainty estimation is through prediction probability from a Softmax
distribution \cite{hendrycks2016baseline}.
However, as the input data gets farther from the training data, this probability estimation naturally yields extrapolations with unsupported high confidence \citep{gal2016dropout}.
Instead, \citet{gal2016dropout} proposed the \textit{Monte Carlo} dropout approach to estimate uncertainty by iteratively applying dropouts to all layers of the model and calculating the variance of generated outputs.
Such estimations based on the probability of a single ground truth label overlooks the many factors that contribute to uncertainty \cite{klas2018uncertainty}. 
In contrast, \citet{passonneau2014benefits} demonstrate the benefits of measuring uncertainty for the ground truth label by fitting a probabilistic model to individual annotators’ observed labels.  
%
Similarly, we demonstrate that calculating annotation disagreement by predicting a set of annotations for the input yields a better estimation of uncertainty than estimations based on the probability of the majority label. 

\label{sec_background}

\section{Methodology}
\begin{figure*}
    \centering
    \includegraphics[width=0.7\textwidth]{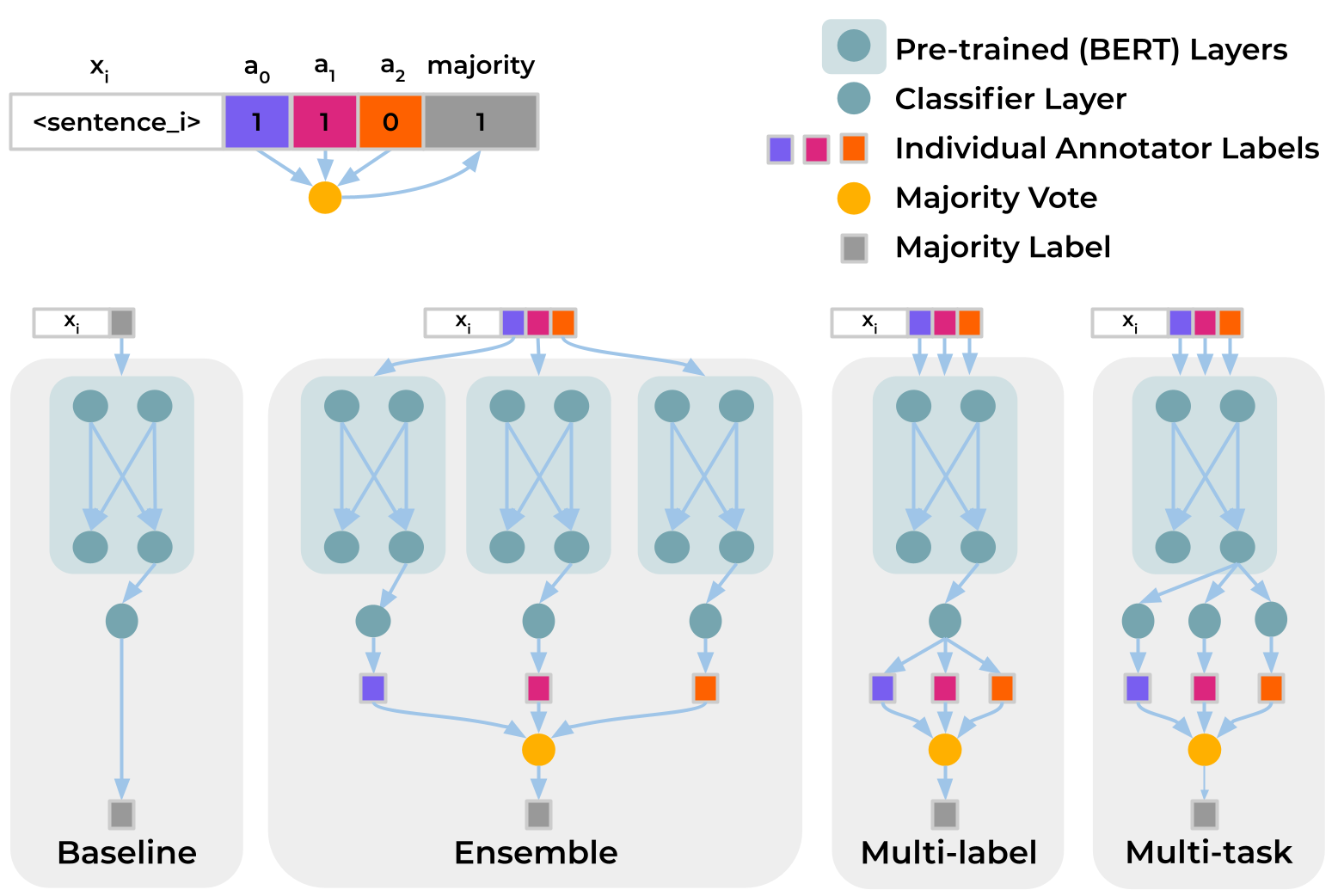}
    \caption{Comparison between approaches for multi-annotator model (ensemble, multi-label and multi-task) and majority label prediction (baseline). Annotation prediction models are trained based on all annotations and apply majority voting to predict the final label.}
    \label{fig:models}
\end{figure*}
 
We define the classification task on an annotated dataset $D = (X, A, Y)$, in which $X$ is a set of text instances, $A$ is the set of annotators and $Y$ is the annotation matrix, in which each entry $y_{ij} \in \{0, 1\}$ represents the label assigned to $x_i \in X$ by $a_j \in A$. 
In most annotated datasets $Y$ includes many missing values, because each annotator only labels a subset of all instances.
We use $\bar{y_{i,}}$ to refer to the 
annotations 
present
for item $x_i$. Similarly, we use $\bar{y_{,j}}$ to refer to the
annotations 
made by
annotator $a_j$.
The classification task aims to predict $maj(\bar{y_{i,}}) \in \{0,1\}$, which is the label assigned to $x_i$ based on the majority vote over $\bar{y_{i,}}$.
We use majority vote, the most commonly used aggregation method; however, our proposed approach leaves open the choice of the aggregation method depending on deployment contexts.

We consider three different multi-annotator architectures: \textit{ensemble}, \textit{multi-label}, and \textit{multi-task}.
Figure~\ref{fig:models} shows the schematic differences between these three variations. 
All variations use Bidirectional Encoder Representations from Transformers \citep[BERT-base;][]{delvim2019bert}.
For each instance $x_i$, a generic representation $h_i \in \mathbb{R}^d$ is generated by the pre-trained BERT-base, and then fine-tuned along with other components of the classifier during training. The size of the representation vector, $d$, is defined by the BERT configuration and is set to 768 for the pre-trained BERT-base. While our experiments are all performed with BERT-base, our methods are not restricted to BERT in their nature, and can be implemented with other pre-trained language models, e.g., RoBERTa \citep{liu2019roberta}.

\vspace{-5px}
\subsection{Baseline model using majority labels}

The baseline model captures the most common approach: a single-task classifier trained to predict the aggregated label for each instance (i.e., majority vote, in our case).
It is built by adding a fully-connected layer to BERT-base outputs ($h_i$). The fully-connected layer applies a linear transformation followed by Softmax function to generate the probability of the majority label, $P(maj(\bar{y_{i,}}) | h_i)$. Compared to the other models described in this section, the baseline model does not make use of annotation matrix $Y$, as it directly predicts the aggregated label $maj(\bar{y_{i,}})$.

\vspace{-5px}
\subsection{Ensemble Approach}

An intuitive approach towards multi-annotator models might be to train an ensemble of models, each trained on different annotators' labels. This approach is not always practical, 
as it may increase the training time prohibitively.
The ensemble approach applies $|A|$ single-task classifiers, each for training and predicting the annotations generated by one annotator. During training, the $j$-th classifier is independently fine-tuned to predict $\bar{y_{,j}}$, which includes all annotations provided by the $j$-th annotator. During test time, we aggregate the outputs by the majority vote of all $|A|$ models to predict $P(maj(\bar{y_{i,}})|x_i)$.\footnote{During prediction, multi-annotator models do not have access to the list of annotators who originally provided the labels for each instance. Therefore, the original majority vote is predicted as the majority vote among all annotators.}

\vspace{-5px}
\subsection{Multi-label Approach}

A more practical approach for multi-annotator modeling
is to consider the problem as a multi-label problem where each label denotes individual annotators' labels.
More specifically, the multi-label approach attempts to learn to predict $|A|$ labels for each input using a multi-label classification framework.
The model first adds a fully-connected layer to transform each $h_i$ to a $|A|$-dimensional vector, and then applies a Sigmoid function to the $j$-th dimension to generate $y_{ij}$. Since $Y$ includes many missing values, the classification loss is calculated based on the available labels $y_{ij} \in \bar{y_{i,}}$. However,  during test time, all $|A|$ outputs are aggregated to predict $P(maj(\bar{y_{i,}})|x_i)$.

\vspace{-5px}
\subsection{Multi-task Approach}

The multi-task based multi-annotator approach attempts to learn multiple annotators' perspectives (labels) as separate classification tasks, all of which share encoder layers to generate the same representation of the input sentence $h_i$, each with its separate fully-connected layer and softmax activation. Compared to the multi-label approach, the multi-task model includes a fully-connected layer explicitly fine-tuned for each annotator. However, compared to the ensemble approach, the representation layers which generate $h_i$ are fine-tuned based on the outputs of all annotation tasks. The loss function is created as the summation of all available labels $\bar{y_{i,}}$ for each instance $x_i$. During test time, the model considers the outputs of all annotation tasks to predict the majority label $P(maj(\bar{y_{i,}}) | x_i)$.

\label{sec_methodology}

\section{Experiments}

\vspace{-5px}
\subsection{Data}
\label{sec_data}

For this study, we  perform experiments on two datasets annotated for subjective tasks: Gab Hate Corpus \cite[GHC;][]{kennedy2020gab} and GoEmotions dataset \cite{demszky2020goemotions}. Both datasets capture per-annotator labels for instances along with corresponding annotators' anonymous ID,
allowing us to model each annotator separately. 

\vspace{-5px}
\subsubsection{Gab Hate Corpus (GHC)} 

GHC \citep{kennedy2020gab}, includes $|X|$ = 27,665 social-media posts collected from a public corpus of Gab.com \citep{pushshift_gab}, each annotated for whether or not they contain hate speech. \citet{kennedy2020gab} define hate speech as language that dehumanizes, attacks human dignity, derogates, incites violence, or supports hateful ideology, such
as white supremacy.
Each instance in GHC is annotated by at least three annotators from a set of $|A|$ = 18 annotators. 
The number of annotations varies for each instance ($M(|\bar{y_{i,}}|)=3.13$ , $SD(|\bar{y_{i,}}|)=0.39$), and in total, there are $86,529$ annotations.
The number of annotated instances per annotator also varies significantly ($M(\bar{y_{,j}})=4807.17$, $SD(\bar{y_{,j}})=3184.89$). 

\vspace{-5px}
\subsubsection{GoEmotions} 
We use a subset of the GoEmotions dataset \citep{demszky2020goemotions} which contains Reddit posts annotated for 28 emotions, split across pre-defined train ($|X|_{train}$ = 43,410), test ($|X|_{test}$ = 5,427) and validation ($|X|_{val}$ = 5,426) subsets. Our experiments focus on the emotion annotations for the six Ekman \cite{ekman1992argument} emotions --- \textit{anger}, \textit{disgust}, \textit{fear}, \textit{joy}, \textit{sadness}, and \textit{surprise}.
Each instance in GoEmotions is annotated by three to five annotators from a set of $|A|$ = 82 annotators.
The number of annotations varies for each instance ($M(|\bar{y_{i,}}|)= 3.58$ , $SD(|\bar{y_{i,}}|)=0.91$), and in total, there are $194,412$ annotations. 
The number of annotated instances varies significantly across annotators ($M(\bar{y_{,j}})=2370.88$, $SD(\bar{y_{,j}})=2180.02$).

\begin{table*}[ht]
    \centering
    \scalebox{0.9}{
    \begin{tabular}{lcccccccc}
        && \multicolumn{3}{c}{Majority Vote} && \multicolumn{3}{c}{Individual Labels} 
        \\\cline{3-5}\cline{7-9}
        Model && Precision & Recall & $\text{F}_1$ && Precision & Recall & $\text{F}_1$
        \\\cline{1-1}\cline{3-5}\cline{7-9}
        Baseline && 49.53$\pm 3.8$ & \textbf{68.78$\pm 4.4$} & 57.32$\pm 1.2$ && - & - & - \\
        Ensemble && 63.98$\pm 1.1$ & 46.09$\pm 1.9$ & 53.54$\pm 1.0$ && 60.92$\pm 0.7$ & 60.97$\pm 0.8$ & 60.94$\pm 0.3$\\
        Multi-label && \textbf{66.02$\pm 2.2$} & 50.16$\pm 2.0$ & 56.94$\pm 1.0$ && \textbf{67.22}$\pm 1.4$ & 55.33$\pm 2.0$ & 60.65$\pm 0.7$\\
        Multi-task && 59.03$\pm 0.9$ & 59.98$\pm 0.6$ & \textbf{59.49$\pm 0.2$} && 63.71$\pm 1.3$ & \textbf{62.76}$\pm 1.5$ & \textbf{63.20}$\pm 0.3$\\
    \end{tabular}    
    }
    \caption{The average and standard deviation of precision, recall, and f-score of model predictions on the \textbf{GHC} dataset, evaluated during 5 iterations of 5-fold stratified cross validation. Majority Vote section represent models' performance on predicting the majority vote, while Individual Labels section reports performance on predicting each raw annotation.
    }
    \label{tab:ghc_prediction_combined}
\end{table*}

\vspace{-5px}
\subsection{Experimental Setup}
We implemented the classification models using the \texttt{transformers} (v3.1) library from HuggingFace \citep{wolf2019huggingface}. 
The training steps employ the \textit{Adam} optimizer \citep{kingma2014adam}. Our experiment settings are configured similar to \citet{kennedy2020gab} and \citet{demszky2020goemotions}, GHC experiments are conducted with a learning rates of $e-7$ and are trained for three epochs, whereas experiments on GoEmotions apply early stopping with a learning rate of $5e-6$.
Since GHC does not have specific train and test subsets, we conducted 5 iterations of stratified 5-fold cross-validations for evaluation, changing only the random state for each iteration. GoEmotions experiments are performed as six different binary classification tasks, also repeated for 5 iterations, using the pre-defined train and test sets. 

\vspace{-5px}
\subsection{Results on GHC}
\label{sec:ghc_res}
\subsubsection{Prediction Results}


Table \ref{tab:ghc_prediction_combined} reports the average and standard deviation of the precision, recall, and $\text{F}_1$-scores for various models, across the 5 iterations. The baseline model, which is trained using the majority vote as ground truth, is also tested against the majority vote labels. For the {ensemble}, {multi-label}, and {multi-task} models, we conduct two types of evaluation: first, we test how well the majority vote of predicted labels match the majority vote of annotations (columns 2-4 in Table \ref{tab:ghc_prediction_combined}); second, we report how well the individual predicted labels for each instance match the annotations (where available) by annotators (columns 5-7 in Table \ref{tab:ghc_prediction_combined}).

We observe that the ensemble model performs significantly worse ($\text{F}_1$=53.54) than the baseline single-task model ($\text{F}_1$=57.32) in predicting majority label. This is presumably due to the fact that each base model in the ensemble is trained using only the examples labeled by the corresponding annotator. Since the number of annotations varies significantly for different annotators (see Section \ref{sec_data}), many base models end up with lower performance, resulting in lower overall performance. 

Multi-label and multi-task models share most layers across different annotator heads. Thus, each annotator head benefits from the updates to the shared layers owing to all instances, regardless of whether they annotated it or not. The multi-label model performs slightly worse ($\text{F}_1$=56.94) than the baseline model. In contrast, the multi-task model, which has a fully connected layer fine-tuned for each annotator, posted a significantly higher F-score ($\text{F}_1$=59.49) than the baseline model. In other words, fine-tuning each annotator head separately and then taking the majority vote performs better than taking the majority vote first and then training on that noisier label. 

Moreover, the baseline model yields higher performance variance among different iterations, such that its standard deviations of precision, recall, and $\text{F}_1$ exceeds those of the other three methods. One possible explanation is that aggregating annotations based on majority votes disposes of information about each annotator and inserts noise into the labels. In other words, modeling each annotator, and their presumable internal consistency, could lead to more stable prediction results. 
However, this hypothesis requires further investigation.

We now evaluate the individual predictions made by the multi-annotator model (prior to majority vote) on how well they match individual annotators' labels (Table \ref{tab:ghc_prediction_combined}). 
All three multi-annotator approaches obtain higher $\text{F}_1$-scores than how the baseline model does in predicting majority labels 
(note that these are different tasks, and not directly comparable).
The multi-task model achieved the highest $\text{F}_1$-score of 63.20. The result suggests that the multi-task model benefits from fine-tuning  annotators separately (thereby avoiding inconsistencies due to majority votes) as well as learning from all instances in a shared fashion. 

\begin{figure}[b]
    \centering
    \includegraphics[width=.46\textwidth]{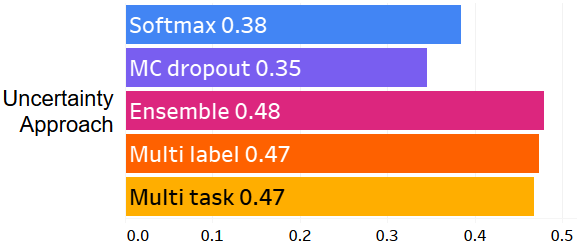}
    \caption{Correlation of different approaches for estimating prediction uncertainty with annotation disagreement on the \textbf{GHC}. Annotation modeling approaches better correlate with disagreement.}
    \label{fig:uncertainty_corr}
\end{figure}

\subsubsection{Modeling Uncertainty}

Next, we study how well we can model uncertainty in predictions. We compare uncertainty in predictions with annotator disagreement, measured as the variance of the annotations. 
\begin{equation}\label{eq:var}
\sigma^2(\bar{y_{i,}}) = \frac{\sum[y_{ij} = 1]\sum[y_{ij} = 0]}{|\bar{y_{i,}}|^2}
\end{equation}
Since the {ensemble}, {multi-label}, and {multi-task} models all make separate predictions corresponding to each annotator, we can calculate the uncertainty in predictions to be the variance of the predicted annotations for each instance $x_i$. 
However, modeling prediction uncertainty in the case of single predictions
is an open question. 
We compare our results with other common approaches for estimating uncertainty in single-task predictions such as \textit{Softmax} probability of the final output for predicting majority vote \citep{hendrycks2016baseline}, and Monte Carlo dropouts \cite{gal2016dropout}, or \textit{MC dropout}, which iteratively applies dropouts to all layers of the model and calculates the variance in predictions. 



Figure \ref{fig:uncertainty_corr} shows the correlations of  uncertainty estimation using each method with the annotation disagreement calculated as $\sigma^2(\bar{y_{i,}})$. 
While traditional estimations such as Softmax and MC dropout have a moderate correlation with annotator disagreements, the uncertainty measured by our three multi-annotator methods show significantly better correlation, with the ensemble method posting a slightly higher correlation than the other two methods. In other words, in addition to performing better on predicting majority votes, multi-annotator models also predict model uncertainty better than traditional approaches.

\begin{figure}
    \centering
    \includegraphics[width=.48\textwidth]{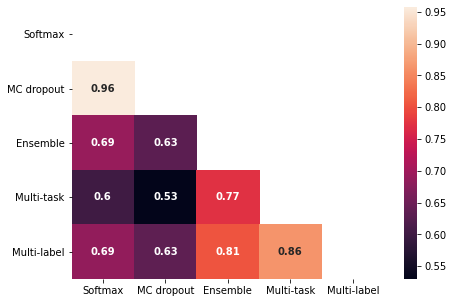}
    \caption{Correlation matrix of approaches for estimating uncertainty. MC dropout and Softmax have high correlation. Our multi-annotator models also have higher internal correlations.
    }
    \label{fig:uncertainty_corr_pairwise}
\end{figure}

We further analyze the pair-wise correlation between estimations of uncertainty by different approaches (Figure~\ref{fig:uncertainty_corr_pairwise}). As expected, the Softmax and MC dropout methods are highly correlated, and similarly, our methods show high correlation among themselves. It is also interesting to note that the uncertainty estimated by our methods also correlate significantly with traditional methods (i.e., between 0.6 and 0.7), except for the multi-task method and MC Dropout method which have a lower correlation of 0.53.

The fact that the uncertainty scores for multi-task and multi-label models are highly correlated with each other (0.86) suggests that they both identify textual features that causes disagreement. We verified this by training a separate model using the same BERT-based setup using Sigmoid activation to directly predict the annotator disagreement. The predicted uncertainty by this model obtained similar correlation with the annotator uncertainty (0.47) as the multi-task and multi-label models.

\subsubsection{Computation Time}
\label{sec:time_res}
We now assess the computation cost  associated with the different approaches.
Table \ref{tab:time} shows the time it took to train a single cross-validation fold, i.e., 80\% of the dataset. As expected, the ensemble approach takes the longest to train, as it require training $|A|$ different models (each with varying training set sizes), and the baseline takes the shortest time. Impressively, multi-label and multi-task models do not take significantly more time to train. In other words, while the multi-task model train additional layers for annotators, it adds only a marginal computation cost to the baseline model.


\begin{table}[t]
    \centering
    \scalebox{0.8}{
    \begin{tabular}{cc}
        Models & Training Time (in mins) \\\hline
        Baseline & 20.5\\
        Ensemble & 158.4 \\ 
        Multi-label & 22.8\\
        Multi-task & 22.3 \\
    \end{tabular}
    }
    \caption{Training time (in minutes); the time it takes to train each model on 80\% of the GHC.}
    \label{tab:time}
\end{table}


\vspace{-5px}
\subsection{Results on GoEmotions}
\label{sec:emo_res}

In this section, we describe results obtained on the six binary classification tasks performed using the GoEmotions dataset. Since the multi-task approach obtained better performance overall on GHC, we report the results on only the multi-task approach here. 
We start by assessing how well the multi-annotator model matches the single-task performance of predicting the majority label.
Table \ref{tab:emotion_prediction_combined} reports the average and standard deviation of $\text{F}_1$-scores over 5 iterations of training and testing. 
Unlike GHC where we used 5-fold cross validation, for the GoEmotions dataset we use the pre-defined train, validation, test splits in the dataset. We verified that these splits are stratifed w.r.t. annotators.
As in GHC experiments, while the baseline model is trained and tested on the majority vote, the multi-task model is trained on available annotator-level annotations for each instance and the predictions from all classifier heads are aggregated to get the final label during testing. 

\begin{table}[b]
    \centering
    \scalebox{0.76}{
    \begin{tabular}{lcccc}
        & \multicolumn{2}{c}{Full Dataset ($|A|$ = 82)}& \multicolumn{2}{c}{Subset ($|A|$ = 53)} 
        \\\hline
        Emotion & Baseline & 
        Multi-task & Baseline & 
        Multi-task 
        \\\hline
        Anger & \textbf{40.38}$\pm4.4$ & 39.01$\pm6.4$ & 41.95$\pm6.1$ & \textbf{42.75}$\pm4.4$ \\
        Disgust & \textbf{38.79}$\pm3.9$& 38.31$\pm1.9$ & \textbf{37.72}$\pm2.0$ & 35.77$\pm2.0$ \\
        Fear & \textbf{58.96}$\pm5.0$& 54.97$\pm6.1$ & 57.68$\pm3.7$ & \textbf{58.58}$\pm2.3$\\
        Joy & 47.80$\pm2.2$ & \textbf{49.53}$\pm3.6$ & \textbf{47.45}$\pm3.1$ & 46.26$\pm1.2$\\
        Sadness & 49.22$\pm5.2$& \textbf{50.36}$\pm3.2$ & 47.55$\pm5.4$ & \textbf{48.00}$\pm3.4$ \\
        Surprise & \textbf{40.96}$\pm2.9$ & 38.97$\pm3.6$ & 39.44$\pm5.7$ & \textbf{40.22}$\pm2.2$ \\
        
        
    \end{tabular}}
    \caption{The average and standard deviation of model prediction f-score on the \textbf{GoEmotions} dataset, evaluated across 5 iterations using the pre-defined train-test splits in the dataset.}
    \label{tab:emotion_prediction_combined}
\end{table}

Results obtained on the full dataset is shown in the second and third columns of Table~\ref{tab:emotion_prediction_combined}. While the multi-task model outperformed the baseline in predicting two emotions --- \textit{joy} and \textit{sadness}, it underperformed the baseline for the other four emotions, although the ranges of $\text{F}_1$-scores largely overlap. It is also observed that the standard deviations of the multi-task model $\text{F}_1$-scores are significantly larger than what was observed for GHC.

On further inspection, we found that many annotators contributed very few annotations in the dataset. For instance, 29 annotators had fewer than 1000 annotations in the training set, six of them having fewer than 100. In addition, the label distribution is extremely skewed for all six emotions --- ranging from 1.6\% positive labels for \textit{fear} on average across all annotators, to 4.0\% positive labels on average for \textit{joy}. Consequently, many annotator heads have too few positive instances to learn from; some had zero positive instances in the training set. This makes the corresponding learning tasks in the multi-task setting hard or even impossible on this dataset, and might explain the lower performance and higher variance in $\text{F}_1$-scores.

In order to make a fairer comparison, we performed our experiments on a subset of the dataset which only includes the annotations by 53 annotators who had more than 1000 annotations. Results obtained on this subset are in the fourth and fifth columns of Table~\ref{tab:emotion_prediction_combined}.
Our multi-annotator model outperforms the baseline model on predicting the majority label in four of the six tasks --- \textit{anger}, \textit{fear}, \textit{sadness}, and \textit{surprise}, while obtaining slightly lower results on \textit{disgust} and \textit{joy}. 
While $\text{F}_1$-score ranges of baseline and multi-task models still largely overlap, the multi-task model fares significantly better when there are enough instances for each annotator head to learn from.
The multi-task model also reported lower standard deviation in performance than the baseline model, suggesting better robustness in the learned model.

\begin{figure}
    \centering
    \includegraphics[width=.49\textwidth]{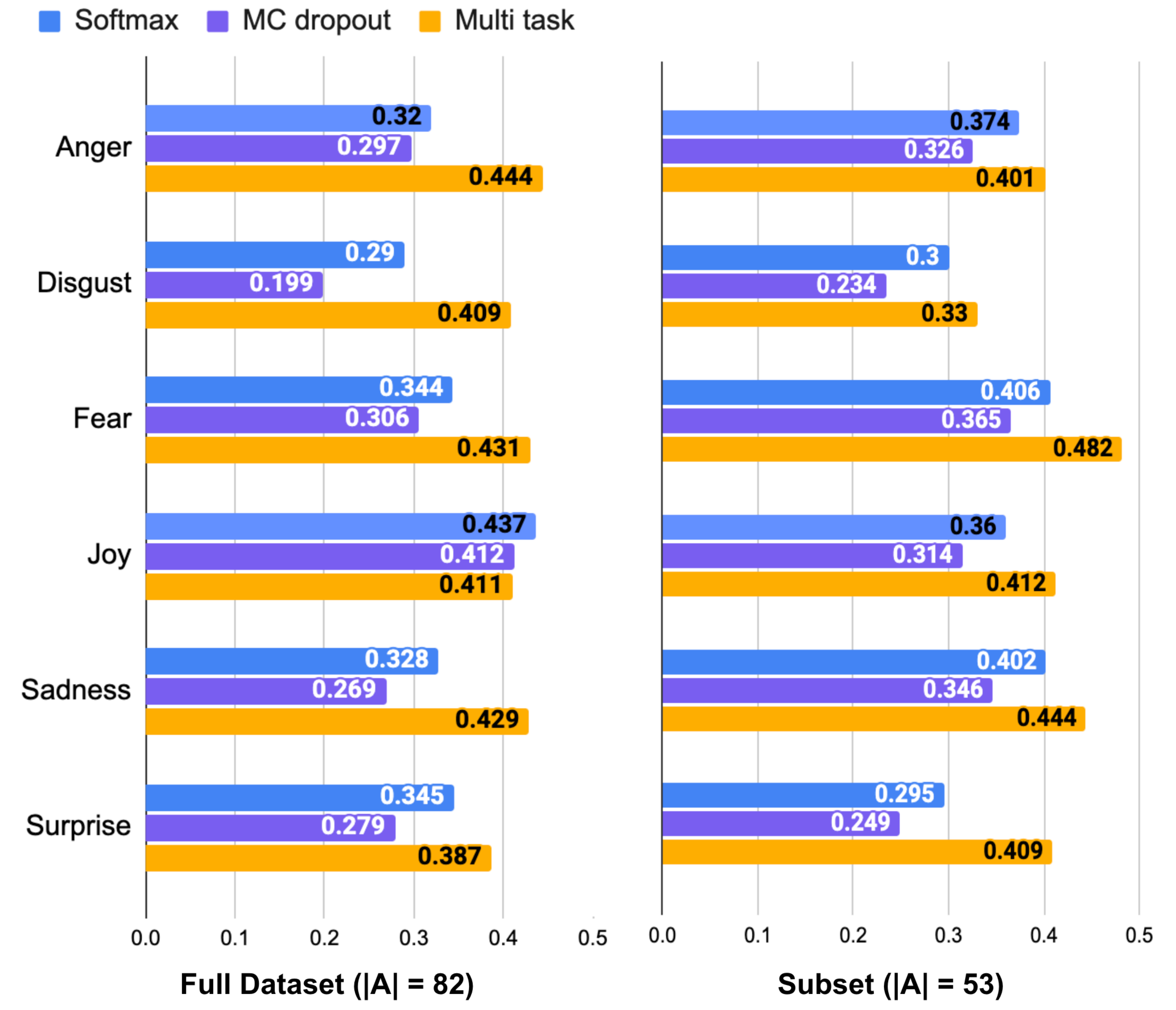}
    \caption{Correlation of different approaches for estimating prediction uncertainty with annotation disagreement for the \textbf{GoEmotions} dataset.}
    \label{fig:emo_uncertainty}
\end{figure}

The main advantage of our multi-annotator model is the ability to capture multiple perspectives efficiently. In that respect, our model fared better at modeling annotator uncertainty across board. 
As shown in Figure \ref{fig:emo_uncertainty}, our multi-annotator model obtained better correlation overall with annotator disagreement than Softmax and MC dropout approaches across all six emotions, both in the full dataset as well as the subset (\textit{joy} in the full dataset being the only exception). 
This further demonstrates the strength of our approach that does not come at any significant cost of performance or efficiency (training the multi-task model on the full dataset takes 6.1 minutes per epoch, comparing to 5.2 minutes for the baseline model).

\label{sec_experiments}

\section{Analysis}

In this section, we further analyze the multi-task model and its outputs, as it posted the overall best performance among the three approaches, considering the predictive  performance, uncertainty modeling correlation, and time efficiency. We focus on the GHC model for this analysis.
\vspace{-5px}
\subsection{Error Analysis}
\label{sec_error}

\begin{table*}[t]
    \centering
    \scalebox{0.8}{
    \begin{tabular}{lccp{.01\textwidth}cccc}
        & \multicolumn{2}{c}{Annotation} && \multicolumn{2}{c}{Multi-task} && Baseline\\\cline{2-3}\cline{5-6}
        Text & Label & \%Hate &&  Label & \%Hate && Label \\\hline
        Agreed. Traitors deserve to be shot. PERIOD. & Not-Hate & 0.33 && Not-Hate & 0.39 && Hate \\
        Are the democrats guilty of treason? & Not-Hate & 0.25 && Not-Hate & 0.39 && Hate \\
        I don’t know, you tell me ya prick. & Not-Hate & 0.33 &&  Not-Hate & 0.39 && Hate \\
        \hline
        They should force the ‘refugees’ to wear this crap. & Hate & 0.67 && Not-Hate & 0.33 && Hate \\
        At this point, nobody gives a shit! & Hate & 0.67 && Not-Hate & 0.44 && Hate \\
        It's OK to be a Nazi!! & Hate & 1.00 && Not-Hate & 0.44 && Hate \\
        \hline
        Arise WHITE PEOPLE! We're back.& Not-Hate & 0.33 && Hate & 0.56 && Not-Hate \\
        Shocking! A jew defending another jew! Who’d have thought? & Not-Hate & 0.33 && Hate & 0.56 && Not-Hate\\
        I am beginning to think being straight is misconduct. & Not-Hate & 0.0 && Hate & 0.56 && Not-Hate \\
        \hline
        Armenia is a nation of mongrel bastards.& Hate & 1.0  && Hate & 0.78 && Not-Hate \\
        Hope they both get AIDS. & Hate & 1.0 && Hate & 0.72 && Not-Hate \\
        I am so NOT afraid of you gay boy. & Hate & 0.67 && Hate & 0.83 && Not-Hate \\

    \end{tabular}
    }
    \caption{Examples from the GHC, for which the baseline differ from multi-task predictions' majority vote.
     (We acknowledge that individual readers may disagree with the annotation labels presented above.)}
    \label{tab:errors}
\end{table*}

We first qualitatively analyze the mismatches between the multi-task and baseline model on their majority vote predictions. Among all GHC instances ($|X| = 27,665$), multi-task and baseline model disagreed on 1,945 labels. Table \ref{tab:errors} shows some examples of such instances and the corresponding majority vote, and the percentage of annotators who labeled them as hate speech. Table \ref{tab:errors} also provides the baseline model's prediction (columns 6), the multi-task model's majority label, and the percentage of prediction heads labeling them as hate speech (columns 4-5). 

The most common type of mismatch (57.94\% of mismatches) occurs when an instance deemed non-hateful (by majority vote of annotations) is correctly labeled by the multi-task model but incorrectly labeled by the baseline (first set of rows in Table \ref{tab:errors}). In other words, these samples represent the baseline model's false-positive predictions, most of which include specific tokens, such as slur words and social group tokens. The next most common type of model mismatch (22.31\% of mismatches) occurred when an instance that was deemed hateful (by majority vote) is mislabeled by the multi-task model and labeled correctly by the baseline model. In general, these two types of mismatches correspond to the positive predictions of the baseline model. A possible explanation for the frequency of such mismatches is the high rate of positive predictions by the baseline model, which is also supported by the higher recall and lower precision scores of the baseline model (Table \ref{tab:ghc_prediction_combined}). 

The other two types of mismatches occurred when the baseline and multi-task model respectively predicted hateful and non-hateful labels. When this mismatch is over an instance deemed hateful by majority vote of annotations (12.19\% of mismatches) the multi-task model is making a false-positive error and we observe mentions of social group names in the text. A large number of such instances had even split (54\% - 44\%) between labels across individual predictions (see Table~\ref{tab:errors}), suggesting the model was unsure. The least common type of disagreement is over instances deemed as hateful by both majority vote of annotations and our multi-task model, but mis-classified by the baseline model (7.56\% of mismatches).

\vspace{-5px}
\subsection{Uncertainty vs. Error}

\begin{figure}[t]
    \centering
    \includegraphics[width=0.48\textwidth]{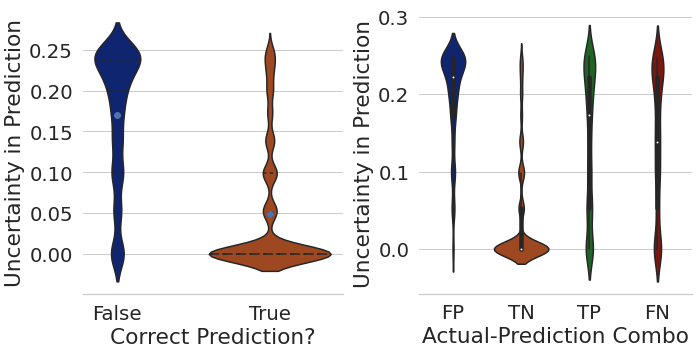}
    \caption{Violin plots denoting distribution across uncertainty for true positive, false positive, false negative, and true negative predictions on GHC. 
    }
    \label{fig:uncertainty_violin}
\end{figure}

Now, we investigate whether the uncertainty in predictions is correlated with whether the multi-task model was able to correctly predict the majority label. 
Note that the value of uncertainty, based on Equation \ref{eq:var}, falls between 0 and 0.25.
We observe that the mean value for uncertainty in correct predictions was 0.049 compared to 0.170 when the model was incorrect. Figure~\ref{fig:uncertainty_violin}a shows the corresponding violin plots. While most incorrect predictions had high uncertainty, a small but significant number of errors were made with certainty.

Separating this analysis across true positives, false positives, false negatives, and true negatives represents a more informative picture. For instance, the model is almost always certain about true negatives ($M$(uncertainty) = 0.040). Similarly, the model is almost always uncertain about false positives ($M$(uncertainty) = 0.199), something we also observed in the error analysis presented in Section~\ref{sec_error}. On the other hand, both true positives and false negatives have a bi-modal distribution of uncertainty, with similar mean uncertainty values of 0.140 and 0.141, respectively. In sum, a negative prediction with high uncertainty is more likely to be a false negative, in our case. 


\label{sec_analysis}

\section{Discussion}
We presented multi-annotator approaches that predict individual labels corresponding with each annotator of a subjective task, as an alternative to the more common practice of deriving (and predicting) a single ``ground-truth'' label, such as the majority vote or average of multiple annotations. We demonstrate that our method based on multi-task architecture obtains better performance for modeling each annotator (63.2 $\text{F}_1$-score, micro-averaged across annotators in GHC), and even when aggregating annotators' predictions, our approach matches or outperforms the baseline across seven tasks.
Our study focuses on majority vote as the baseline aggregation approach to demonstrate how this commonly used approach loses meaningful information. Other aggregation strategies such as MACE \citep{hovy2013learning} and Bayesian methods \citep{paun2018comparing} could be explored in future work as complementary approaches that can work with the multi-annotator framework.

\vspace{-5px}
\subsection{Advantages of Multi-Annotator Modeling}
One core advantage of our method, which can further be leveraged in practice, is its ability to provide multiple predictions for each instance. As demonstrated in Figure~\ref{fig:uncertainty_corr} and \ref{fig:emo_uncertainty}, the multiple predictions can derive an uncertainty estimation that better matches with the disagreement between annotators. The estimated uncertainty could be used to determine \textit{when not to make a prediction} or to route the example to a manual content moderation queue as it may be an example that annotators likely disagreed on. One could also investigate how to learn an uncertainty threshold to make cleverer predictions. For instance, based on our analysis in \ref{sec_analysis}, a negative prediction with high uncertainty is very likely to be a false negative. One could use this knowledge in a deployment scenario and predict a positive label in case of a negative majority prediction with high uncertainty. 

Predicting multiple annotations rather than a ground truth is specifically essential in subjective tasks. As \citet{alm2008affect} argues, in many subjective tasks, the aim is not to find an \textit{accurate} answer; instead, a model can produce the most \textit{acceptable} answer based on responses from different judgements. Accordingly, our method contrasts with approaches for enhancing ground-truth generation prior to modeling. Our approach aims to preserve annotators' consistency in labeling by delaying the annotation aggregation until the final stage. As a final step, if required, application-driven approaches can be employed to find the most proper answer. For instance, an aggregation approach based on MACE \citep{hovy2013learning, paun2018comparing}, could be applied to the predicted individual labels to find a final label that considers the trustworthiness of individual annotators. 

Researchers have pointed out that in more objective tasks, such as commonsense knowledge or word sense disambiguation, training a model on judgements of a specific set of annotators lack generalizability to annotations generated by new annotators \citep{geva2019we}. However, in subjective tasks such as affect and online abuse detection, different annotator perspectives, and their contrasts can be useful
\citep{gordon2021disagreement}. 

Another advantage of having multiple prediction heads in a multi-task architecture is that we could adapt the same model to different value systems. For instance, in cases where annotators with different moral beliefs systemically produce different labels \cite{waseem2016you,diaz2020biases,patton2019annotating}, one could use the multi-task approach to have a single global model that can adjust predictions to be conditioned on different value systems. This is valuable for international media platforms to build and deploy global models that attend to local cultures and values without retraining entirely separate models for each culture.

Multi-annotator modeling can also be applied in scenarios that may benefit from obtaining several perspectives for a single instance. For example, in detecting affect in language, a range of subjective human knowledge, interpretation, and experience can be modeled through a multi-annotator architecture. This approach would generate a range of affective states either along affect categories, such as anger and happiness, or dimensions, such as arousal and pleasantness \cite{alm2011subjective, alm2008affect}, which correspond with different subjective perceptions of the text. Another example is sarcasm detection, where an ambiguous sarcastic text is labeled differently according to annotators' thresholds for sarcasm \citep{rakov2013sure}. In a multi-annotator setting, internal consistency of each annotators' threshold for sarcasm may be preserved in the training process.

\vspace{-5px}
\subsection{Limitations and Challenges}
Our approach is not without limitations. Our experiments were computationally viable because of the relatively small number of annotators in our annotator pool (18 for GHC and 82 for the GoEmotions dataset), which is not usually the case with large crowd-sourced datasets. For instance, the dataset by \citet{diaz2020biases} has over 1.4K individual annotators, and \citet{Jigsaw-bias} built a dataset with over 8K annotators. Fine-tuning that many separate annotator heads will be computationally expensive and may not be a viable option. However, clustering annotators based on their agreements and aggregating annotator labels into cluster labels could address this issue. In that scenario, the multi-task model would include separate classifier heads for each cluster of annotators. The number of clusters could be determined based on availability of computational resources and data factors to enhance the multi-task approach. This is an important direction of research for future work.

The proposed approach along with other methods for incorporating individual annotators and their disagreements are only viable when annotated datasets include annotator-level labels for each instance.
However, most multiply annotated datasets contain only per-instance majority labels \cite{waseem2016hateful, Jigsaw-toxic}, or aggregate percentages \cite{davidson2017automated,Jigsaw-bias}.
Even in cases where the raw annotations were released,
the multi-annotator model requires there being enough annotations from each annotator to model them effectively. However, 
we observed that the dataset designers may not have envisioned such a utility of annotator-level labels for downstream analysis. For instance, in the GoEmotions dataset, many annotators labeled fewer than 1000 instances, making it hard for annotator-level modeling.  Moreover, the high cost of gathering large number of annotations per annotator in crowdsourcing platforms may limit the data collection and call for post-hoc modeling solutions.
One way to tackle this issue is by choosing a subset of top-performing annotator heads (during the validation step) for the final prediction. Future work should look into such post-processing steps that could further improve the performance.

To enable further exploration into open questions in studying annotator disagreements and efficient ways to model them, the main challenge is the lack of annotator-level labels. This largely stems from the practice of considering crowd annotators as interchangeable, and not accounting for the differences in their perspectives. We recommend data providers to consider releasing individual annotation labels, when feasible to do so, in an anonymized way and with appropriate consent. We also encourage researchers to design data collection efforts in a way that includes a sufficient number of annotations by each annotator, so that systematic differences in their annotation behaviors could be better understood and accounted for.

\label{sec_discussion}

\section{Conclusion}
We present a multi-annotator approach that employs a different classifier head for each annotator of a dataset as an alternate method to the practice of predicting the aggregated majority vote. We demonstrate that our method can efficiently obtain better performance in modeling each annotator as well as match the majority vote prediction performance. We present experiments across different subjective classification tasks, including hate speech detection and six different emotion detection tasks. The model uncertainty estimated based on our multi-annotator model(s)' predictions obtain a higher correlation to the annotation disagreement than more traditional methods. We expect future works to investigate our multi-annotator approach as a means to detect and mitigate model biases. Moreover, monitoring the performance of annotator heads and model uncertainty in an active learning setting has the potential to capture a more diverse and comprehensive set of perspectives in data. 

\label{sec_conclusion}

\section{Ethical Considerations}

Our paper discusses an approach for attending to individual annotator's judgements in training a supervised model. In doing that, our multi-annotator approach better preserves minority perspectives that are usually sidelined by majority votes. Our intended use case for this approach is in subjective NLP tasks, such as identifying affect, abusive language, or hate speech, where generating a single true answer does not capture the nuances.

While our method likely preserves minority perspectives, 
a misuse of this technique might happen upon weighting individual annotator's labels during prediction. Such an alternation aimed solely to improve the majority label prediction performance may adversely impact the representation of different perspectives in the model. In fact, such an optimization may cause further marginalization to under-represented perspectives than the current majority vote based approaches. For instance, identifying annotator heads that significantly disagree with the majority vote might cause their perspectives being at higher risk of being excluded.

It is also important to consider the number of annotators in the annotator pool when applying this method, in order to protect the privacy and anonymity of annotators, since our approach attempts to model their personal subjective preferences and biases. This is especially critical in the case of sensitive tasks such as hate speech annotations, where associating individual annotators with such representations may be undesirable.

\label{sec_ethical}

\section*{Acknowledgments}
We thank Ben Hutchinson, Lucas Dixon, and Jeffrey Sorensen for their valuable feedback on the manuscript. We also thank TACL action editor, Dirk Hovy, and the anonymous reviewers for their constructive feedback.

\bibliography{anthology,main}

\begin{thebibliography}{91}
\expandafter\ifx\csname natexlab\endcsname\relax\def\natexlab#1{#1}\fi

\bibitem[{Alm(2011)}]{alm2011subjective}
Cecilia~Ovesdotter Alm. 2011.
\newblock \href {https://aclanthology.org/P11-2019.pdf} {Subjective natural
  language problems: Motivations, applications, characterizations, and
  implications}.
\newblock In \emph{Proceedings of the 49th Annual Meeting of the Association
  for Computational Linguistics: Human Language Technologies}, pages 107--112.

\bibitem[{Alm(2008)}]{alm2008affect}
Ebba Cecilia~Ovesdotter Alm. 2008.
\newblock \href
  {http://citeseerx.ist.psu.edu/viewdoc/download?doi=10.1.1.172.9934&rep=rep1&type=pdf}
  {\emph{Affect in* Text and Speech}}.
\newblock Ph.D. thesis, University of Illinois at Urbana-Champaign.

\bibitem[{Alonso et~al.(2015)Alonso, Johannsen, de~Lacalle, and
  Agirre}]{alonso2015predicting}
H{\'e}ctor~Mart{\'\i}nez Alonso, Anders Johannsen, Oier~Lopez de~Lacalle, and
  Eneko Agirre. 2015.
\newblock \href {https://aclanthology.org/W15-2711.pdf} {Predicting word sense
  annotation agreement}.
\newblock In \emph{Proceedings of the First Workshop on Linking Computational
  Models of Lexical, Sentential and Discourse-level Semantics}, pages 89--94.

\bibitem[{Aman and Szpakowicz(2007)}]{aman2007identifying}
Saima Aman and Stan Szpakowicz. 2007.
\newblock \href
  {https://citeseerx.ist.psu.edu/viewdoc/download?doi=10.1.1.1081.5218&rep=rep1&type=pdf}
  {Identifying expressions of emotion in text}.
\newblock In \emph{International Conference on Text, Speech and Dialogue},
  pages 196--205. Springer.

\bibitem[{Ando et~al.(2018)Ando, Kobashikawa, Kamiyama, Masumura, Ijima, and
  Aono}]{ando2018soft}
Atsushi Ando, Satoshi Kobashikawa, Hosana Kamiyama, Ryo Masumura, Yusuke Ijima,
  and Yushi Aono. 2018.
\newblock \href {https://ieeexplore.ieee.org/abstract/document/8461299}
  {Soft-target training with ambiguous emotional utterances for dnn-based
  speech emotion classification}.
\newblock In \emph{2018 IEEE International Conference on Acoustics, Speech and
  Signal Processing (ICASSP)}, pages 4964--4968. IEEE.

\bibitem[{Aroyo and Welty(2013)}]{aroyo2013crowd}
Lora Aroyo and Chris Welty. 2013.
\newblock \href
  {https://d1wqtxts1xzle7.cloudfront.net/66797651/Crowd_Truth_Harnessing_disagreement_in_c20210503-14654-7wfrq6-with-cover-page-v2.pdf?Expires=1627861060&Signature=EmAeg8SzrcVD12SuuXGCe0AhncNMIoGZD1Fi3s9TlJciXzV3mjQ8byE4Ms4n-A-uGehS~hj5Im~NgT7FJ7jOAXPn4~lfQE~zLAAAJoGVCTkIKWwto-eBV2sDmqodhUVummP7K-XTJwDZgeigqzMmSRAllsX~R~Uy8BDcO7OYv6Y5hXWgfcZMMzj6i~pqJG-rF1UC-Pdlr3LzFsObyup1TRAiAdbdWhQ3plle-FFu9UZzPV3WhLqK6jqdFzc77JDScux1tJE53nijB3Plx4G5o750yjO9TfcO6FbZVy1FzcQrob-dw-SAdHre4PaZ5U0-fTKltnbAGQO09sk3Tc8Nkw__&Key-Pair-Id=APKAJLOHF5GGSLRBV4ZA}
  {Crowd truth: Harnessing disagreement in crowdsourcing a relation extraction
  gold standard}.
\newblock \emph{WebSci2013. ACM}, 2013.

\bibitem[{Breitfeller et~al.(2019)Breitfeller, Ahn, Jurgens, and
  Tsvetkov}]{breitfeller2019finding}
Luke Breitfeller, Emily Ahn, David Jurgens, and Yulia Tsvetkov. 2019.
\newblock \href {https://aclanthology.org/D19-1176.pdf} {Finding
  microaggressions in the wild: A case for locating elusive phenomena in social
  media posts}.
\newblock In \emph{Proceedings of the 2019 Conference on Empirical Methods in
  Natural Language Processing and the 9th International Joint Conference on
  Natural Language Processing (EMNLP-IJCNLP)}, pages 1664--1674.

\bibitem[{Buechel and Hahn(2017)}]{buechel2017emobank}
Sven Buechel and Udo Hahn. 2017.
\newblock \href {https://aclanthology.org/E17-2092.pdf} {Emobank: Studying the
  impact of annotation perspective and representation format on dimensional
  emotion analysis}.
\newblock In \emph{Proceedings of the 15th Conference of the European Chapter
  of the Association for Computational Linguistics: Volume 2, Short Papers},
  pages 578--585.

\bibitem[{Chandrasekharan et~al.(2019)Chandrasekharan, Gandhi, Mustelier, and
  Gilbert}]{chandrasekharan2019crossmod}
Eshwar Chandrasekharan, Chaitrali Gandhi, Matthew~Wortley Mustelier, and Eric
  Gilbert. 2019.
\newblock Crossmod: A cross-community learning-based system to assist reddit
  moderators.
\newblock \emph{Proceedings of the ACM on human-computer interaction},
  3(CSCW):1--30.

\bibitem[{Cheplygina and Pluim(2018)}]{cheplygina2018crowd}
Veronika Cheplygina and Josien~PW Pluim. 2018.
\newblock \href {https://arxiv.org/pdf/1806.08174.pdf} {Crowd disagreement
  about medical images is informative}.
\newblock In \emph{Intravascular imaging and computer assisted stenting and
  large-scale annotation of biomedical data and expert label synthesis}, pages
  105--111. Springer.

\bibitem[{Chou and Lee(2019)}]{chou2019every}
Huang-Cheng Chou and Chi-Chun Lee. 2019.
\newblock \href
  {https://www.researchgate.net/profile/Huang-Cheng-Chou/publication/332791139_Every_Rating_Matters_Joint_Learning_of_Subjective_Labels_and_Individual_Annotators_for_Speech_Emotion_Classification/links/5df0d69ba6fdcc283717cca3/Every-Rating-Matters-Joint-Learning-of-Subjective-Labels-and-Individual-Annotators-for-Speech-Emotion-Classification.pdf}
  {Every rating matters: Joint learning of subjective labels and individual
  annotators for speech emotion classification}.
\newblock In \emph{ICASSP 2019-2019 IEEE International Conference on Acoustics,
  Speech and Signal Processing (ICASSP)}, pages 5886--5890. IEEE.

\bibitem[{Cohn and Specia(2013)}]{cohn-specia-2013-modelling}
Trevor Cohn and Lucia Specia. 2013.
\newblock \href {https://www.aclweb.org/anthology/P13-1004} {Modelling
  annotator bias with multi-task {G}aussian processes: An application to
  machine translation quality estimation}.
\newblock In \emph{Proceedings of the 51st Annual Meeting of the Association
  for Computational Linguistics (Volume 1: Long Papers)}, pages 32--42, Sofia,
  Bulgaria. Association for Computational Linguistics.

\bibitem[{Corazza et~al.(2020)Corazza, Menini, Cabrio, Tonelli, and
  Villata}]{corazza2020multilingual}
Michele Corazza, Stefano Menini, Elena Cabrio, Sara Tonelli, and Serena
  Villata. 2020.
\newblock \href {https://hal.archives-ouvertes.fr/hal-02972184/document} {A
  multilingual evaluation for online hate speech detection}.
\newblock \emph{ACM Transactions on Internet Technology (TOIT)}, 20(2):1--22.

\bibitem[{Cowan and Khatchadourian(2003)}]{cowan2003empathy}
Gloria Cowan and D{\'e}sir{\'e}e Khatchadourian. 2003.
\newblock \href
  {https://citeseerx.ist.psu.edu/viewdoc/download?doi=10.1.1.852.3266&rep=rep1&type=pdf}
  {Empathy, ways of knowing, and interdependence as mediators of gender
  differences in attitudes toward hate speech and freedom of speech}.
\newblock \emph{Psychology of Women Quarterly}, 27(4):300--308.

\bibitem[{Cowen et~al.(2019)Cowen, Sauter, Tracy, and
  Keltner}]{cowen2019mapping}
Alan Cowen, Disa Sauter, Jessica~L Tracy, and Dacher Keltner. 2019.
\newblock \href {https://journals.sagepub.com/doi/pdf/10.1177/1529100619850176}
  {Mapping the passions: Toward a high-dimensional taxonomy of emotional
  experience and expression}.
\newblock \emph{Psychological Science in the Public Interest}, 20(1):69--90.

\bibitem[{Crowdflower(2016)}]{crowdflower}
Crowdflower. 2016.
\newblock
  \url{https://www.figureeight.com/data/sentiment-analysis-emotion-text/}.

\bibitem[{Davidson et~al.(2019)Davidson, Bhattacharya, and
  Weber}]{davidson2019racial}
Thomas Davidson, Debasmita Bhattacharya, and Ingmar Weber. 2019.
\newblock \href {https://doi.org/10.18653/v1/W19-3504} {Racial bias in hate
  speech and abusive language detection datasets}.
\newblock In \emph{Proceedings of the Third Workshop on Abusive Language
  Online}, pages 25--35, Florence, Italy. Association for Computational
  Linguistics.

\bibitem[{Davidson et~al.(2017)Davidson, Warmsley, Macy, and
  Weber}]{davidson2017automated}
Thomas Davidson, Dana Warmsley, Michael Macy, and Ingmar Weber. 2017.
\newblock \href
  {https://ojs.aaai.org/index.php/ICWSM/article/download/14955/14805}
  {Automated hate speech detection and the problem of offensive language}.
\newblock In \emph{Proceedings of the International AAAI Conference on Web and
  Social Media}, volume~11.

\bibitem[{Dawid and Skene(1979)}]{dawid1979maximum}
Alexander~Philip Dawid and Allan~M Skene. 1979.
\newblock \href
  {https://citeseerx.ist.psu.edu/viewdoc/download?doi=10.1.1.469.1377&rep=rep1&type=pdf}
  {Maximum likelihood estimation of observer error-rates using the em
  algorithm}.
\newblock \emph{Journal of the Royal Statistical Society: Series C (Applied
  Statistics)}, 28(1):20--28.

\bibitem[{De~Marneffe et~al.(2012)De~Marneffe, Manning, and Potts}]{de2012did}
Marie-Catherine De~Marneffe, Christopher~D Manning, and Christopher Potts.
  2012.
\newblock \href
  {https://direct.mit.edu/coli/article-pdf/38/2/301/1801598/coli_a_00097.pdf}
  {Did it happen? the pragmatic complexity of veridicality assessment}.
\newblock \emph{Computational linguistics}, 38(2):301--333.

\bibitem[{Demszky et~al.(2020)Demszky, Movshovitz-Attias, Ko, Cowen, Nemade,
  and Ravi}]{demszky2020goemotions}
Dorottya Demszky, Dana Movshovitz-Attias, Jeongwoo Ko, Alan Cowen, Gaurav
  Nemade, and Sujith Ravi. 2020.
\newblock \href {https://aclanthology.org/2020.acl-main.372} {{G}o{E}motions: A
  dataset of fine-grained emotions}.
\newblock In \emph{Proceedings of the 58th Annual Meeting of the Association
  for Computational Linguistics}, Online. Association for Computational
  Linguistics.

\bibitem[{Desmet and Hoste(2013)}]{desmet2013emotion}
Bart Desmet and V{\'e}ronique Hoste. 2013.
\newblock \href
  {https://www.researchgate.net/profile/Bart-Desmet/publication/257405079_Emotion_detection_in_suicide_notes/links/5d92229e92851c33e94b24a6/Emotion-detection-in-suicide-notes.pdf}
  {Emotion detection in suicide notes}.
\newblock \emph{Expert Systems with Applications}, 40(16):6351--6358.

\bibitem[{Devlin et~al.(2019)Devlin, Chang, Lee, and
  Toutanova}]{delvim2019bert}
J.~Devlin, Ming-Wei Chang, Kenton Lee, and Kristina Toutanova. 2019.
\newblock \href
  {https://arxiv.org/pdf/1810.04805.pdf&usg=ALkJrhhzxlCL6yTht2BRmH9atgvKFxHsxQ}
  {Bert: Pre-training of deep bidirectional transformers for language
  understanding}.
\newblock In \emph{NAACL-HLT}.

\bibitem[{D{\'\i}az(2020)}]{diaz2020biases}
Mark D{\'\i}az. 2020.
\newblock \href
  {http://markjdiaz.com/wp-content/uploads/2021/01/Diaz_BiasesAsValues.pdf}
  {\emph{Biases as Values: Evaluating Algorithms in Context}}.
\newblock Ph.D. thesis, Northwestern University.

\bibitem[{D{\'\i}az et~al.(2018)D{\'\i}az, Johnson, Lazar, Piper, and
  Gergle}]{diaz2018addressing}
Mark D{\'\i}az, Isaac Johnson, Amanda Lazar, Anne~Marie Piper, and Darren
  Gergle. 2018.
\newblock \href {https://dl.acm.org/doi/pdf/10.1145/3173574.3173986}
  {Addressing age-related bias in sentiment analysis}.
\newblock In \emph{Proceedings of the 2018 CHI Conference on Human Factors in
  Computing Systems}, pages 1--14.

\bibitem[{Dixon et~al.(2018)Dixon, Li, Sorensen, Thain, and
  Vasserman}]{dixon2018measuring}
Lucas Dixon, John Li, Jeffrey Sorensen, Nithum Thain, and Lucy Vasserman. 2018.
\newblock \href {https://dl.acm.org/doi/pdf/10.1145/3278721.3278729} {Measuring
  and mitigating unintended bias in text classification}.
\newblock In \emph{Proceedings of the 2018 AAAI/ACM Conference on AI, Ethics,
  and Society}, pages 67--73.

\bibitem[{Dumitrache(2015)}]{dumitrache2015crowdsourcing}
Anca Dumitrache. 2015.
\newblock \href
  {https://link.springer.com/chapter/10.1007/978-3-319-18818-8_43}
  {Crowdsourcing disagreement for collecting semantic annotation}.
\newblock In \emph{European Semantic Web Conference}, pages 701--710. Springer.

\bibitem[{Ekman(1992)}]{ekman1992argument}
Paul Ekman. 1992.
\newblock \href {https://asset-pdf.scinapse.io/prod/1966797434/1966797434.pdf}
  {An argument for basic emotions}.
\newblock \emph{Cognition \& emotion}, 6(3-4):169--200.

\bibitem[{Fayek et~al.(2016)Fayek, Lech, and Cavedon}]{fayek2016modeling}
Haytham~M Fayek, Margaret Lech, and Lawrence Cavedon. 2016.
\newblock \href {https://ieeexplore.ieee.org/abstract/document/7727250/}
  {Modeling subjectiveness in emotion recognition with deep neural networks:
  Ensembles vs soft labels}.
\newblock In \emph{2016 international joint conference on neural networks
  (IJCNN)}, pages 566--570. IEEE.

\bibitem[{Fornaciari et~al.(2021)Fornaciari, Uma, Paun, Plank, Hovy, and
  Poesio}]{fornaciari-etal-2021-beyond}
Tommaso Fornaciari, Alexandra Uma, Silviu Paun, Barbara Plank, Dirk Hovy, and
  Massimo Poesio. 2021.
\newblock \href {https://doi.org/10.18653/v1/2021.naacl-main.204} {Beyond black
  {\&} white: Leveraging annotator disagreement via soft-label multi-task
  learning}.
\newblock In \emph{Proceedings of the 2021 Conference of the North American
  Chapter of the Association for Computational Linguistics: Human Language
  Technologies}, pages 2591--2597, Online. Association for Computational
  Linguistics.

\bibitem[{Gaffney(2018)}]{pushshift_gab}
Gavin Gaffney. 2018.
\newblock Pushshift gab corpus.
\newblock \url{https://files.pushshift.io/gab/}.
\newblock Accessed: 2019-5-23.

\bibitem[{Gal and Ghahramani(2016)}]{gal2016dropout}
Yarin Gal and Zoubin Ghahramani. 2016.
\newblock \href {http://proceedings.mlr.press/v48/gal16.pdf} {Dropout as a
  bayesian approximation: Representing model uncertainty in deep learning}.
\newblock In \emph{international conference on machine learning}, pages
  1050--1059. PMLR.

\bibitem[{Geva et~al.(2019)Geva, Goldberg, and Berant}]{geva2019we}
Mor Geva, Yoav Goldberg, and Jonathan Berant. 2019.
\newblock \href {https://doi.org/10.18653/v1/D19-1107} {Are we modeling the
  task or the annotator? an investigation of annotator bias in natural language
  understanding datasets}.
\newblock In \emph{Proceedings of the 2019 Conference on Empirical Methods in
  Natural Language Processing and the 9th International Joint Conference on
  Natural Language Processing (EMNLP-IJCNLP)}, pages 1161--1166, Hong Kong,
  China. Association for Computational Linguistics.

\bibitem[{Ghandeharioun et~al.(2019)Ghandeharioun, Eoff, Jou, and
  Picard}]{ghandeharioun2019characterizing}
Asma Ghandeharioun, Brian Eoff, Brendan Jou, and Rosalind Picard. 2019.
\newblock \href {https://arxiv.org/pdf/1909.09285.pdf} {Characterizing sources
  of uncertainty to proxy calibration and disambiguate annotator and data
  bias}.
\newblock In \emph{2019 IEEE/CVF International Conference on Computer Vision
  Workshop (ICCVW)}, pages 4202--4206. IEEE.

\bibitem[{Gordon et~al.(2021)Gordon, Zhou, Patel, Hashimoto, and
  Bernstein}]{gordon2021disagreement}
Mitchell~L Gordon, Kaitlyn Zhou, Kayur Patel, Tatsunori Hashimoto, and
  Michael~S Bernstein. 2021.
\newblock \href {http://www.kayur.org/papers/chi2021.pdf} {The disagreement
  deconvolution: Bringing machine learning performance metrics in line with
  reality}.
\newblock In \emph{Proceedings of the 2021 CHI Conference on Human Factors in
  Computing Systems}.

\bibitem[{Greifeneder et~al.(2017)Greifeneder, Bless, and
  Fiedler}]{greifeneder2017social}
Rainer Greifeneder, Herbert Bless, and Klaus Fiedler. 2017.
\newblock \href
  {https://www.taylorfrancis.com/books/mono/10.4324/9781315784731/social-cognition-herbert-bless-klaus-fiedler}
  {\emph{Social cognition: How individuals construct social reality}}.
\newblock Psychology Press.

\bibitem[{Hendrycks and Gimpel(2017)}]{hendrycks2016baseline}
Dan Hendrycks and Kevin Gimpel. 2017.
\newblock \href {https://arxiv.org/pdf/1610.02136.pdf} {A baseline for
  detecting misclassified and out-of-distribution examples in neural networks}.
\newblock \emph{Proceedings of International Conference on Learning
  Representations}.

\bibitem[{Hirschberg et~al.(2003)Hirschberg, Liscombe, and
  Venditti}]{hirschberg2003experiments}
Julia Hirschberg, Jackson Liscombe, and Jennifer Venditti. 2003.
\newblock \href
  {https://www.researchgate.net/profile/Julia-Hirschberg/publication/246140762_Experiments_in_Emotional_Speech/links/556f094208aeab7772282916/Experiments-in-Emotional-Speech.pdf}
  {Experiments in emotional speech}.
\newblock In \emph{ISCA \& IEEE Workshop on Spontaneous Speech Processing and
  Recognition}.

\bibitem[{Hirschberg and Manning(2015)}]{hirschberg2015advances}
Julia Hirschberg and Christopher~D Manning. 2015.
\newblock \href
  {https://nlp.stanford.edu/~manning/xyzzy/Hirschberg-Manning-Science-2015.pdf}
  {Advances in natural language processing}.
\newblock \emph{Science}, 349(6245):261--266.

\bibitem[{Hovy et~al.(2013)Hovy, Berg-Kirkpatrick, Vaswani, and
  Hovy}]{hovy2013learning}
Dirk Hovy, Taylor Berg-Kirkpatrick, Ashish Vaswani, and Eduard Hovy. 2013.
\newblock \href {https://aclanthology.org/N13-1132.pdf} {Learning whom to trust
  with mace}.
\newblock In \emph{Proceedings of the 2013 Conference of the North American
  Chapter of the Association for Computational Linguistics: Human Language
  Technologies}, pages 1120--1130.

\bibitem[{Hutchinson et~al.(2020)Hutchinson, Prabhakaran, Denton, Webster,
  Zhong, and Denuyl}]{hutchinson2020social}
Ben Hutchinson, Vinodkumar Prabhakaran, Emily Denton, Kellie Webster, Yu~Zhong,
  and Stephen Denuyl. 2020.
\newblock \href {https://doi.org/10.18653/v1/2020.acl-main.487} {Social biases
  in {NLP} models as barriers for persons with disabilities}.
\newblock In \emph{Proceedings of the 58th Annual Meeting of the Association
  for Computational Linguistics}, pages 5491--5501, Online. Association for
  Computational Linguistics.

\bibitem[{Jigsaw(2018)}]{Jigsaw-toxic}
Jigsaw. 2018.
\newblock \href
  {https://www.kaggle.com/c/\\jigsaw-toxic-comment-classification-challenge/data}
  {Toxic comment classification challenge}.
\newblock Accessed: 2021-05-01.

\bibitem[{Jigsaw(2019)}]{Jigsaw-bias}
Jigsaw. 2019.
\newblock \href
  {https://www.kaggle.com/c/\\jigsaw-unintended-bias-in-toxicity-classification/data}
  {Unintended bias in toxicity classification}.
\newblock Accessed: 2021-05-01.

\bibitem[{Jurgens et~al.(2019)Jurgens, Hemphill, and
  Chandrasekharan}]{jurgens-etal-2019-just}
David Jurgens, Libby Hemphill, and Eshwar Chandrasekharan. 2019.
\newblock \href {https://doi.org/10.18653/v1/P19-1357} {A just and
  comprehensive strategy for using {NLP} to address online abuse}.
\newblock In \emph{Proceedings of the 57th Annual Meeting of the Association
  for Computational Linguistics}, pages 3658--3666, Florence, Italy.
  Association for Computational Linguistics.

\bibitem[{Kairam and Heer(2016)}]{kairam2016parting}
Sanjay Kairam and Jeffrey Heer. 2016.
\newblock \href {https://dl.acm.org/doi/pdf/10.1145/2818048.2820016} {Parting
  crowds: Characterizing divergent interpretations in crowdsourced annotation
  tasks}.
\newblock In \emph{Proceedings of the 19th ACM Conference on Computer-Supported
  Cooperative Work \& Social Computing}, pages 1637--1648.

\bibitem[{Kennedy et~al.(2020)Kennedy, Atari, Mostafazadeh~Davani, Yeh, Omrani,
  Kim, Coombs~Jr., Havaldar, Portillo-Wightman, Gonzalez, Hoover, Azatian,
  Cardenas, Hussain, Lara, Omary, Park, Wang, Wijaya, Zhang, Meyerowitz, and
  Dehghani}]{kennedy2020gab}
Brendan Kennedy, Mohammad Atari, Aida Mostafazadeh~Davani, Leigh Yeh, Ali
  Omrani, Yehsong Kim, Kris Coombs~Jr., Shreya Havaldar, Gwenyth
  Portillo-Wightman, Elaine Gonzalez, Joe Hoover, Aida Azatian, Gabriel
  Cardenas, Alyzeh Hussain, Austin Lara, Adam Omary, Christina Park, Xin Wang,
  Clarisa Wijaya, Yong Zhang, Beth Meyerowitz, and Morteza Dehghani. 2020.
\newblock \href {https://doi.org/10.31234/osf.io/hqjxn} {The gab hate corpus: A
  collection of 27k posts annotated for hate speech}.

\bibitem[{Kingma and Ba(2015)}]{kingma2014adam}
Diederik~P. Kingma and Jimmy Ba. 2015.
\newblock \href {http://arxiv.org/abs/1412.6980} {Adam: {A} method for
  stochastic optimization}.
\newblock In \emph{3rd International Conference on Learning Representations,
  {ICLR} 2015, San Diego, CA, USA, May 7-9, 2015, Conference Track
  Proceedings}.

\bibitem[{Kl{\"a}s and Vollmer(2018)}]{klas2018uncertainty}
Michael Kl{\"a}s and Anna~Maria Vollmer. 2018.
\newblock \href
  {https://link.springer.com/chapter/10.1007/978-3-319-99229-7_36} {Uncertainty
  in machine learning applications: A practice-driven classification of
  uncertainty}.
\newblock In \emph{International Conference on Computer Safety, Reliability,
  and Security}, pages 431--438. Springer.

\bibitem[{Krippendorff(2011)}]{krippendorff2011agreement}
Klaus Krippendorff. 2011.
\newblock \href
  {https://repository.upenn.edu/cgi/viewcontent.cgi?article=1286&context=asc_papers}
  {Agreement and information in the reliability of coding}.
\newblock \emph{Communication Methods and Measures}, 5(2):93--112.

\bibitem[{Liscombe et~al.(2003)Liscombe, Venditti, and
  Hirschberg}]{liscombe2003classifying}
Jackson Liscombe, Jennifer Venditti, and Julia Hirschberg. 2003.
\newblock \href
  {https://academiccommons.columbia.edu/doi/10.7916/D8VX0QTJ/download}
  {Classifying subject ratings of emotional speech using acoustic features}.
\newblock In \emph{Eighth European Conference on Speech Communication and
  Technology}.

\bibitem[{Liu et~al.(2010)}]{liu2010sentiment}
Bing Liu et~al. 2010.
\newblock Sentiment analysis and subjectivity.
\newblock \emph{Handbook of natural language processing}, 2(2010):627--666.

\bibitem[{Liu et~al.(2003)Liu, Lieberman, and Selker}]{liu2003model}
Hugo Liu, Henry Lieberman, and Ted Selker. 2003.
\newblock \href
  {https://citeseerx.ist.psu.edu/viewdoc/download?doi=10.1.1.73.2973&rep=rep1&type=pdf}
  {A model of textual affect sensing using real-world knowledge}.
\newblock In \emph{Proceedings of the 8th international conference on
  Intelligent user interfaces}, pages 125--132.

\bibitem[{Liu(2020)}]{liu2020human}
Tong Liu. 2020.
\newblock \href
  {https://scholarworks.rit.edu/cgi/viewcontent.cgi?article=11619&context=theses}
  {Human-in-the-loop learning from crowdsourcing and social media}.

\bibitem[{Liu et~al.(2019)Liu, He, Chen, and Gao}]{liu2019multi}
Xiaodong Liu, Pengcheng He, Weizhu Chen, and Jianfeng Gao. 2019.
\newblock \href {https://doi.org/10.18653/v1/P19-1441} {Multi-task deep neural
  networks for natural language understanding}.
\newblock In \emph{Proceedings of the 57th Annual Meeting of the Association
  for Computational Linguistics}, pages 4487--4496, Florence, Italy.
  Association for Computational Linguistics.

\bibitem[{Luo et~al.(2020)Luo, Card, and Jurafsky}]{luo-etal-2020-detecting}
Yiwei Luo, Dallas Card, and Dan Jurafsky. 2020.
\newblock \href {https://doi.org/10.18653/v1/2020.findings-emnlp.296}
  {Detecting stance in media on global warming}.
\newblock In \emph{Findings of the Association for Computational Linguistics:
  EMNLP 2020}, pages 3296--3315, Online. Association for Computational
  Linguistics.

\bibitem[{Mihalcea and Liu(2006)}]{mihalcea2006corpus}
Rada Mihalcea and Hugo Liu. 2006.
\newblock \href
  {https://www.aaai.org/Papers/Symposia/Spring/2006/SS-06-03/SS06-03-027.pdf}
  {A corpus-based approach to finding happiness.}
\newblock In \emph{AAAI Spring Symposium: Computational Approaches to Analyzing
  Weblogs}, pages 139--144.

\bibitem[{Mishra et~al.(2019)Mishra, Yannakoudakis, and
  Shutova}]{mishra2019tackling}
Pushkar Mishra, Helen Yannakoudakis, and Ekaterina Shutova. 2019.
\newblock \href {https://arxiv.org/pdf/1908.06024.pdf} {Tackling online abuse:
  A survey of automated abuse detection methods}.
\newblock \emph{arXiv preprint arXiv:1908.06024}.

\bibitem[{Mower et~al.(2009)Mower, Metallinou, Lee, Kazemzadeh, Busso, Lee, and
  Narayanan}]{mower2009interpreting}
Emily Mower, Angeliki Metallinou, Chi-Chun Lee, Abe Kazemzadeh, Carlos Busso,
  Sungbok Lee, and Shrikanth Narayanan. 2009.
\newblock \href {http://web.eecs.umich.edu/~emilykmp/EmilyPapers/MowerACII.pdf}
  {Interpreting ambiguous emotional expressions}.
\newblock In \emph{2009 3rd International Conference on Affective Computing and
  Intelligent Interaction and Workshops}, pages 1--8. IEEE.

\bibitem[{Mozafari et~al.(2019)Mozafari, Farahbakhsh, and
  Crespi}]{mozafari2019bert}
Marzieh Mozafari, Reza Farahbakhsh, and Noel Crespi. 2019.
\newblock \href {https://arxiv.org/pdf/1910.12574.pdf} {A bert-based transfer
  learning approach for hate speech detection in online social media}.
\newblock In \emph{International Conference on Complex Networks and Their
  Applications}, pages 928--940. Springer.

\bibitem[{Nowak and R{\"u}ger(2010)}]{nowak2010reliable}
Stefanie Nowak and Stefan R{\"u}ger. 2010.
\newblock \href {https://oro.open.ac.uk/25874/1/mir354s-nowak.pdf} {How
  reliable are annotations via crowdsourcing: a study about inter-annotator
  agreement for multi-label image annotation}.
\newblock In \emph{Proceedings of the international conference on Multimedia
  information retrieval}, pages 557--566.

\bibitem[{Pang and Lee(2004)}]{pang-lee-2004-sentimental}
Bo~Pang and Lillian Lee. 2004.
\newblock \href {https://doi.org/10.3115/1218955.1218990} {A sentimental
  education: Sentiment analysis using subjectivity summarization based on
  minimum cuts}.
\newblock In \emph{Proceedings of the 42nd Annual Meeting of the Association
  for Computational Linguistics ({ACL}-04)}, pages 271--278, Barcelona, Spain.

\bibitem[{Passonneau and Carpenter(2014)}]{passonneau2014benefits}
Rebecca~J Passonneau and Bob Carpenter. 2014.
\newblock \href
  {https://direct.mit.edu/tacl/article-pdf/doi/10.1162/tacl_a_00185/1566907/tacl_a_00185.pdf}
  {The benefits of a model of annotation}.
\newblock \emph{Transactions of the Association for Computational Linguistics},
  2:311--326.

\bibitem[{Patton et~al.(2019)Patton, Blandfort, Frey, Gaskell, and
  Karaman}]{patton2019annotating}
Desmond Patton, Philipp Blandfort, William Frey, Michael Gaskell, and Svebor
  Karaman. 2019.
\newblock \href
  {https://scholarspace.manoa.hawaii.edu/bitstream/10125/59653/0213.pdf}
  {Annotating social media data from vulnerable populations: Evaluating
  disagreement between domain experts and graduate student annotators}.
\newblock In \emph{Proceedings of the 52nd Hawaii International Conference on
  System Sciences}.

\bibitem[{Paun et~al.(2018)Paun, Carpenter, Chamberlain, Hovy, Kruschwitz, and
  Poesio}]{paun2018comparing}
Silviu Paun, Bob Carpenter, Jon Chamberlain, Dirk Hovy, Udo Kruschwitz, and
  Massimo Poesio. 2018.
\newblock \href
  {https://direct.mit.edu/tacl/article-pdf/doi/10.1162/tacl_a_00040/1567662/tacl_a_00040.pdf}
  {Comparing bayesian models of annotation}.
\newblock \emph{Transactions of the Association for Computational Linguistics},
  6:571--585.

\bibitem[{Plank et~al.(2014)Plank, Hovy, and
  S{\o}gaard}]{plank-etal-2014-learning}
Barbara Plank, Dirk Hovy, and Anders S{\o}gaard. 2014.
\newblock \href {https://doi.org/10.3115/v1/E14-1078} {Learning part-of-speech
  taggers with inter-annotator agreement loss}.
\newblock In \emph{Proceedings of the 14th Conference of the {E}uropean Chapter
  of the Association for Computational Linguistics}, pages 742--751,
  Gothenburg, Sweden. Association for Computational Linguistics.

\bibitem[{Plutchik(1980)}]{plutchik1980general}
Robert Plutchik. 1980.
\newblock \href
  {https://www.sciencedirect.com/science/article/pii/B9780125587013500077} {A
  general psychoevolutionary theory of emotion}.
\newblock In \emph{Theories of emotion}, pages 3--33. Elsevier.

\bibitem[{Poria et~al.(2019)Poria, Majumder, Mihalcea, and
  Hovy}]{poria2019emotion}
Soujanya Poria, Navonil Majumder, Rada Mihalcea, and Eduard Hovy. 2019.
\newblock \href {https://ieeexplore.ieee.org/stamp/stamp.jsp?arnumber=8764449}
  {Emotion recognition in conversation: Research challenges, datasets, and
  recent advances}.
\newblock \emph{IEEE Access}, 7:100943--100953.

\bibitem[{Prabhakaran et~al.(2012)Prabhakaran, Bloodgood, Diab, Dorr, Levin,
  Piatko, Rambow, and Van~Durme}]{prabhakaran-etal-2012-statistical}
Vinodkumar Prabhakaran, Michael Bloodgood, Mona Diab, Bonnie Dorr, Lori Levin,
  Christine~D. Piatko, Owen Rambow, and Benjamin Van~Durme. 2012.
\newblock \href {https://www.aclweb.org/anthology/W12-3807} {Statistical
  modality tagging from rule-based annotations and crowdsourcing}.
\newblock In \emph{Proceedings of the Workshop on Extra-Propositional Aspects
  of Meaning in Computational Linguistics}, pages 57--64, Jeju, Republic of
  Korea. Association for Computational Linguistics.

\bibitem[{Prabhakaran et~al.(2021)Prabhakaran, Davani, and
  Diaz}]{prabhakaran2021releasing}
Vinodkumar Prabhakaran, Aida~Mostafazadeh Davani, and Mark Diaz. 2021.
\newblock On releasing annotator-level labels and information in datasets.
\newblock In \emph{Proceedings of the 15th Linguistic Annotation Workshop},
  Virtual. Association for Computational Linguistics.

\bibitem[{Prabhakaran et~al.(2019)Prabhakaran, Hutchinson, and
  Mitchell}]{prabhakaran-etal-2019-perturbation}
Vinodkumar Prabhakaran, Ben Hutchinson, and Margaret Mitchell. 2019.
\newblock \href {https://doi.org/10.18653/v1/D19-1578} {Perturbation
  sensitivity analysis to detect unintended model biases}.
\newblock In \emph{Proceedings of the 2019 Conference on Empirical Methods in
  Natural Language Processing and the 9th International Joint Conference on
  Natural Language Processing (EMNLP-IJCNLP)}, pages 5740--5745, Hong Kong,
  China. Association for Computational Linguistics.

\bibitem[{Prabhakaran et~al.(2020)Prabhakaran, Waseem, Akiwowo, and
  Vidgen}]{prabhakaran-etal-2020-online}
Vinodkumar Prabhakaran, Zeerak Waseem, Seyi Akiwowo, and Bertie Vidgen. 2020.
\newblock \href {https://doi.org/10.18653/v1/2020.alw-1.1} {Online abuse and
  human rights: {WOAH} satellite session at {R}ights{C}on 2020}.
\newblock In \emph{Proceedings of the Fourth Workshop on Online Abuse and
  Harms}, pages 1--6, Online. Association for Computational Linguistics.

\bibitem[{Price et~al.(2020)Price, Gifford-Moore, Flemming, Musker, Roichman,
  Sylvain, Thain, Dixon, and Sorensen}]{price2020six}
Ilan Price, Jordan Gifford-Moore, Jory Flemming, Saul Musker, Maayan Roichman,
  Guillaume Sylvain, Nithum Thain, Lucas Dixon, and Jeffrey Sorensen. 2020.
\newblock \href {https://doi.org/10.18653/v1/2020.alw-1.15} {Six attributes of
  unhealthy conversations}.
\newblock In \emph{Proceedings of the Fourth Workshop on Online Abuse and
  Harms}, pages 114--124, Online. Association for Computational Linguistics.

\bibitem[{Rakov and Rosenberg(2013)}]{rakov2013sure}
Rachel Rakov and Andrew Rosenberg. 2013.
\newblock \href {https://lpp.ilpga.fr/PDF/IS130109/IS130109.PDF} {``sure, i did
  the right thing'': a system for sarcasm detection in speech.}
\newblock In \emph{Interspeech}, pages 842--846.

\bibitem[{Ross et~al.(2010)Ross, Irani, Silberman, Zaldivar, and
  Tomlinson}]{ross2010crowdworkers}
Joel Ross, Lilly Irani, M~Six Silberman, Andrew Zaldivar, and Bill Tomlinson.
  2010.
\newblock \href
  {https://www.academia.edu/download/43592369/Who_are_the_crowdworkers_shifting_demogr20160310-18708-cv9zu3.pdf}
  {Who are the crowdworkers? shifting demographics in mechanical turk}.
\newblock In \emph{CHI'10 extended abstracts on Human factors in computing
  systems}, pages 2863--2872.

\bibitem[{Russell(2003)}]{russell2003core}
James~A Russell. 2003.
\newblock \href {https://www.academia.edu/download/30925178/psyc-rev2003.pdf}
  {Core affect and the psychological construction of emotion.}
\newblock \emph{Psychological review}, 110(1):145.

\bibitem[{Sabou et~al.(2014)Sabou, Bontcheva, Derczynski, and
  Scharl}]{sabou2014corpus}
Marta Sabou, Kalina Bontcheva, Leon Derczynski, and Arno Scharl. 2014.
\newblock \href
  {https://citeseerx.ist.psu.edu/viewdoc/download?doi=10.1.1.1048.7024&rep=rep1&type=pdf}
  {Corpus annotation through crowdsourcing: Towards best practice guidelines.}
\newblock In \emph{LREC}, pages 859--866.

\bibitem[{Sap et~al.(2019)Sap, Card, Gabriel, Choi, and Smith}]{sap2019risk}
Maarten Sap, Dallas Card, Saadia Gabriel, Yejin Choi, and Noah~A Smith. 2019.
\newblock \href {https://www.aclweb.org/anthology/P19-1163.pdf} {The risk of
  racial bias in hate speech detection}.
\newblock In \emph{Proceedings of the 57th annual meeting of the association
  for computational linguistics}, pages 1668--1678.

\bibitem[{Schmidt and Wiegand(2017)}]{schmidt2017survey}
Anna Schmidt and Michael Wiegand. 2017.
\newblock \href {https://www.aclweb.org/anthology/W17-1101.pdf} {A survey on
  hate speech detection using natural language processing}.
\newblock In \emph{Proceedings of the fifth international workshop on natural
  language processing for social media}, pages 1--10.

\bibitem[{Schwab and Karlen(2019)}]{schwab2019cxplain}
Patrick Schwab and Walter Karlen. 2019.
\newblock \href {https://arxiv.org/pdf/1910.12336} {{CXPlain: Causal
  Explanations for Model Interpretation under Uncertainty}}.
\newblock In \emph{{Advances in Neural Information Processing Systems
  (NeurIPS)}}.

\bibitem[{Snow et~al.(2008)Snow, O{'}Connor, Jurafsky, and
  Ng}]{snow-etal-2008-cheap}
Rion Snow, Brendan O{'}Connor, Daniel Jurafsky, and Andrew Ng. 2008.
\newblock \href {https://www.aclweb.org/anthology/D08-1027} {Cheap and fast
  {--} but is it good? evaluating non-expert annotations for natural language
  tasks}.
\newblock In \emph{Proceedings of the 2008 Conference on Empirical Methods in
  Natural Language Processing}, pages 254--263, Honolulu, Hawaii. Association
  for Computational Linguistics.

\bibitem[{Strapparava and Mihalcea(2007)}]{strapparava2007semeval}
Carlo Strapparava and Rada Mihalcea. 2007.
\newblock \href {https://www.aclweb.org/anthology/S07-1013.pdf} {Semeval-2007
  task 14: Affective text}.
\newblock In \emph{Proceedings of the Fourth International Workshop on Semantic
  Evaluations (SemEval-2007)}, pages 70--74.

\bibitem[{Vidgen et~al.(2021)Vidgen, Thrush, Waseem, and
  Kiela}]{vidgen2020learning}
Bertie Vidgen, Tristan Thrush, Zeerak Waseem, and Douwe Kiela. 2021.
\newblock \href {https://doi.org/10.18653/v1/2021.acl-long.132} {Learning from
  the worst: Dynamically generated datasets to improve online hate detection}.
\newblock In \emph{Proceedings of the 59th Annual Meeting of the Association
  for Computational Linguistics and the 11th International Joint Conference on
  Natural Language Processing (Volume 1: Long Papers)}, pages 1667--1682,
  Online. Association for Computational Linguistics.

\bibitem[{Warner and Hirschberg(2012)}]{warner2012detecting}
William Warner and Julia Hirschberg. 2012.
\newblock \href {https://www.aclweb.org/anthology/W12-2103.pdf} {Detecting hate
  speech on the world wide web}.
\newblock In \emph{Proceedings of the second workshop on language in social
  media}, pages 19--26.

\bibitem[{Waseem(2016)}]{waseem2016you}
Zeerak Waseem. 2016.
\newblock \href {https://www.aclweb.org/anthology/W16-5618.pdf} {Are you a
  racist or am i seeing things? annotator influence on hate speech detection on
  twitter}.
\newblock In \emph{Proceedings of the first workshop on NLP and computational
  social science}, pages 138--142.

\bibitem[{Waseem et~al.(2017)Waseem, Davidson, Warmsley, and
  Weber}]{waseem2017understanding}
Zeerak Waseem, Thomas Davidson, Dana Warmsley, and Ingmar Weber. 2017.
\newblock \href {https://arxiv.org/pdf/1705.09899} {Understanding abuse: A
  typology of abusive language detection subtasks}.
\newblock \emph{arXiv preprint arXiv:1705.09899}.

\bibitem[{Waseem and Hovy(2016)}]{waseem2016hateful}
Zeerak Waseem and Dirk Hovy. 2016.
\newblock \href {https://www.aclweb.org/anthology/N16-2013.pdf} {Hateful
  symbols or hateful people? predictive features for hate speech detection on
  twitter}.
\newblock In \emph{Proceedings of the NAACL student research workshop}, pages
  88--93.

\bibitem[{Waterhouse(2013)}]{waterhouse2013pay}
Tamsyn~P Waterhouse. 2013.
\newblock \href {https://research.google/pubs/pub40700.pdf} {Pay by the bit: an
  information-theoretic metric for collective human judgment}.
\newblock In \emph{Proceedings of the 2013 conference on Computer supported
  cooperative work}, pages 623--638.

\bibitem[{Wiebe et~al.(2004)Wiebe, Wilson, Bruce, Bell, and
  Martin}]{wiebe2004learning}
Janyce Wiebe, Theresa Wilson, Rebecca Bruce, Matthew Bell, and Melanie Martin.
  2004.
\newblock \href
  {https://direct.mit.edu/coli/article-pdf/30/3/277/1798072/0891201041850885.pdf}
  {Learning subjective language}.
\newblock \emph{Computational linguistics}, 30(3):277--308.

\bibitem[{Wolf et~al.(2020)Wolf, Debut, Sanh, Chaumond, Delangue, Moi, Cistac,
  Rault, Louf, Funtowicz, Davison, Shleifer, von Platen, Ma, Jernite, Plu, Xu,
  Le~Scao, Gugger, Drame, Lhoest, and Rush}]{wolf2019huggingface}
Thomas Wolf, Lysandre Debut, Victor Sanh, Julien Chaumond, Clement Delangue,
  Anthony Moi, Pierric Cistac, Tim Rault, Remi Louf, Morgan Funtowicz, Joe
  Davison, Sam Shleifer, Patrick von Platen, Clara Ma, Yacine Jernite, Julien
  Plu, Canwen Xu, Teven Le~Scao, Sylvain Gugger, Mariama Drame, Quentin Lhoest,
  and Alexander Rush. 2020.
\newblock \href {https://doi.org/10.18653/v1/2020.emnlp-demos.6} {Transformers:
  State-of-the-art natural language processing}.
\newblock In \emph{Proceedings of the 2020 Conference on Empirical Methods in
  Natural Language Processing: System Demonstrations}, pages 38--45, Online.
  Association for Computational Linguistics.

\bibitem[{Zhou et~al.(2021)Zhou, Sap, Swayamdipta, Choi, and
  Smith}]{zhou2021challenges}
Xuhui Zhou, Maarten Sap, Swabha Swayamdipta, Yejin Choi, and Noah Smith. 2021.
\newblock \href {https://aclanthology.org/2021.eacl-main.274} {Challenges in
  automated debiasing for toxic language detection}.
\newblock pages 3143--3155.

\bibitem[{Zhu et~al.(2020)Zhu, Song, Jin, and Jiang}]{liu2019roberta}
Minghao Zhu, Youzhe Song, Ge~Jin, and Keyuan Jiang. 2020.
\newblock \href {https://doi.org/10.18653/v1/2020.louhi-1.14} {Identifying
  personal experience tweets of medication effects using pre-trained
  {R}o{BERT}a language model and its updating}.
\newblock In \emph{Proceedings of the 11th International Workshop on Health
  Text Mining and Information Analysis}, pages 127--137, Online. Association
  for Computational Linguistics.

\end{thebibliography}
\bibliographystyle{acl_natbib}

\end{document}